\documentclass[letterpaper]{article} % DO NOT CHANGE THIS
\usepackage[preprint]{aaai2027}  % DO NOT CHANGE THIS
\usepackage[hyphens]{url}  % DO NOT CHANGE THIS
\usepackage{graphicx} % DO NOT CHANGE THIS
\usepackage{natbib}  % DO NOT CHANGE THIS AND DO NOT ADD ANY OPTIONS TO IT
\usepackage{caption} % DO NOT CHANGE THIS AND DO NOT ADD ANY OPTIONS TO IT
\usepackage{algorithm}
\usepackage{algorithmic}

\usepackage{newfloat}
\usepackage{listings}
\DeclareCaptionStyle{ruled}{labelfont=normalfont,labelsep=colon,strut=off} % DO NOT CHANGE THIS
\floatstyle{ruled}
\newfloat{listing}{tb}{lst}{}
\floatname{listing}{Listing}

\usepackage{booktabs}

\usepackage{multirow}
\usepackage{colortbl}
\usepackage{amsmath}
\usepackage{amssymb}
\newcommand{\meanstd}[2]{\ensuremath{#1{\pm}#2}}

\definecolor{second}{RGB}{252,228,232}
\definecolor{best}{RGB}{228,239,250}

\title{Agentic Reinforcement Learning with Observation-Calibrated Self-Distillation}

\author{
Yi Yang\textsuperscript{\rm 1,2 *},
Cong Qin\textsuperscript{\rm 1,3,*},
Xiaodan Liu\textsuperscript{\rm 1,4,*},
Chishui Chen\textsuperscript{\rm 1,5,*},
Qing Dong\textsuperscript{\rm 1,6,*},
Yan Zhang\textsuperscript{\rm 1,*},\\
Cao Liu\textsuperscript{\rm 1},
Zhao Yang\textsuperscript{\rm 1},
Lu Pan\textsuperscript{\rm 1},
Jiaye Lin\textsuperscript{\rm 1,$\dagger$},
Yi Feng\textsuperscript{\rm 2,$\dagger$}
}

\affiliations{
\textsuperscript{\rm 1}Meituan LongCat Interaction
\textsuperscript{\rm 2}Nanjing University\\
\textsuperscript{\rm 3}Peking University
\textsuperscript{\rm 4}McMaster University
\textsuperscript{\rm 5}Fudan University
\textsuperscript{\rm 6}East China Normal University
}

\begin{document}
\maketitle

\begingroup
\renewcommand{\thefootnote}{}

\makeatletter
\renewcommand{\@makefntext}[1]{\noindent #1}
\makeatother

\footnotetext{%
\textsuperscript{*}\,Equal contribution.
\textsuperscript{$\dagger$}\,Corresponding authors.\\
Email: yi.yang@smail.nju.edu.cn.\\
Work done during an internship at Meituan.
}
\endgroup

% \begin{abstract}
% Large language model agents are commonly trained with sparse trajectory-level rewards, which provide limited guidance for individual reasoning and action tokens. On-Policy Self-Distillation offers dense supervision by re-scoring generated tokens under a privileged replay view, but the resulting support may conflate future observations with replay-format and action-scaffold effects. We propose Observation-Calibrated Self-Distillation (OCSD), which compares structurally matched Full and Observation-Ablated replay views to derive an observation residual. OCSD applies this residual at high-uncertainty steps to modulate token-level GRPO updates while preserving the trajectory-level update direction. Experiments on ALFWorld, WebShop, and Search-QA across three Qwen3 model scales show consistent gains, including 8.9--14.0 point improvements over GRPO on ALFWorld. Diagnostic analyses further show that the calibrated residual better aligns with local environment feedback.
% \end{abstract}

\begin{abstract}
Large language model agents are commonly trained through reinforcement learning with sparse trajectory-level rewards, which offer limited guidance on how strongly individual tokens should be updated. On-Policy Self-Distillation (OPSD) addresses this by re-scoring generated tokens under a privileged replay view to obtain dense, token-level supervision. However, we identify a confounding issue: the resulting support may reflect both the privileged information contained in the replay view and score shifts induced by the replay scaffold, making it difficult to attribute the support specifically to that information. This issue is especially pronounced when future environment observations serve as privileged information, since replaying them requires reconstructing an extended scaffold that itself perturbs token scores. To resolve this confounding, we propose \textbf{Observation-Calibrated Self-Distillation (OCSD)}, which contrasts two structurally matched replay views, Full and Observation-Ablated, differing only in whether the actual future observation is present, to derive an observation residual that discounts score changes
shared by the replay scaffold. OCSD then applies this residual to modulate token-level GRPO updates at high-uncertainty steps, while preserving the trajectory-level update direction. Experiments on ALFWorld, WebShop, and Search-QA across three Qwen3 model scales show that OCSD consistently outperforms strong baselines. Diagnostic analyses further confirm that the calibrated residual aligns better with local environment feedback. Our code is publicly available at \url{https://github.com/yiy1x/OCSD}.

\end{abstract}

\section{Introduction}
\label{sec:introdcution}
% Large Language Model (LLM) agents are increasingly used for long-horizon interactive tasks such as web navigation~\cite{webarena}, embodied control~\cite{openvla}, and information seeking~\cite{infogent}, which require repeated observation, reasoning, and action~\cite{react,agentbench}. However, reinforcement learning methods such as Group Relative Policy Optimization (GRPO)~\cite{grpo} typically rely on sparse trajectory-level rewards, offering little guidance on how strongly reasoning or action tokens should be updated~\cite{gigpo,turnlevelreward}.

Large Language Model (LLM) agents are increasingly used for long-horizon interactive tasks such as web navigation~\cite{webarena}, embodied control~\cite{openvla}, and information seeking~\cite{infogent}, where completing a task requires many rounds of observation, reasoning, and action~\cite{react,agentbench}. Reinforcement learning has become the dominant paradigm for training such agents, with methods such as Group Relative Policy Optimization (GRPO)~\cite{grpo}. However, GRPO and its variants typically apply a shared trajectory-level reward signal to all tokens in a multi-step trajectory, offering little guidance on how strongly individual tokens generated at each step should be updated~\cite{gigpo,turnlevelreward,actfocus}.

% On-Policy Self-Distillation (OPSD) provides a natural way to complement sparse trajectory-level rewards with fine-grained supervision. It first generates on-policy outputs and then re-evaluates the generated tokens under a privileged replay view after rollout~\citep{opsd}. Recent agentic self-distillation methods further construct local supervision from subsequent environment observations~\citep{hero,serl}, completed interaction trajectories~\citep{stepopsd,rlsd,sdar}, or distilled experience summaries~\citep{sdar,seed}. Although these methods differ in how they construct, localize, and integrate privileged supervision, they commonly use the probability changes induced by the privileged replay view to measure the support provided by the target hindsight information. However, this interpretation implicitly assumes that the additional replay context beyond the target information does not substantially affect token scores.

On-Policy Self-Distillation (OPSD) offers a natural way to complement sparse trajectory-level rewards with fine-grained supervision, by generating on-policy outputs and then re-evaluating the generated tokens under a privileged replay view after rollout~\citep{opsd}. Recent agentic self-distillation methods build such privileged views from subsequent environment observations~\citep{hero,serl}, completed interaction trajectories~\citep{stepopsd,rlsd,sdar}, or distilled experience summaries~\citep{sdar,seed}. Despite differing in how privileged supervision is constructed, localized, and integrated, these methods commonly treat the probability change induced by the replay view as the support contributed by the target hindsight information, implicitly assuming that the added replay context beyond this information leaves token scores otherwise unaffected.

% This issue becomes particularly evident when future environment observations are used as privileged information. As shown in Figure~\ref{fig:motivation}, the agent incorrectly generates ``take the mug from shelf 1,'' after which the environment reports ''there is no mug on the shelf''. The Full replay view incorporates this future observation into the context and re-scores the original output. However, it also introduces an additional replay format and future-action scaffold. Even after the true observation is removed, the Observation-Ablated view still assigns support similar to that of the Full view for some tokens, showing that these changes are not entirely attributable to the environment feedback. Therefore, Full-view support conflates changes associated with the future observation and those shared by the replay scaffold. Controlling for the latter requires a structurally matched view that preserves the same replay structure while excluding the actual future observation.

\begin{figure*}[htbp]
    \centering
    \includegraphics[width=\textwidth]{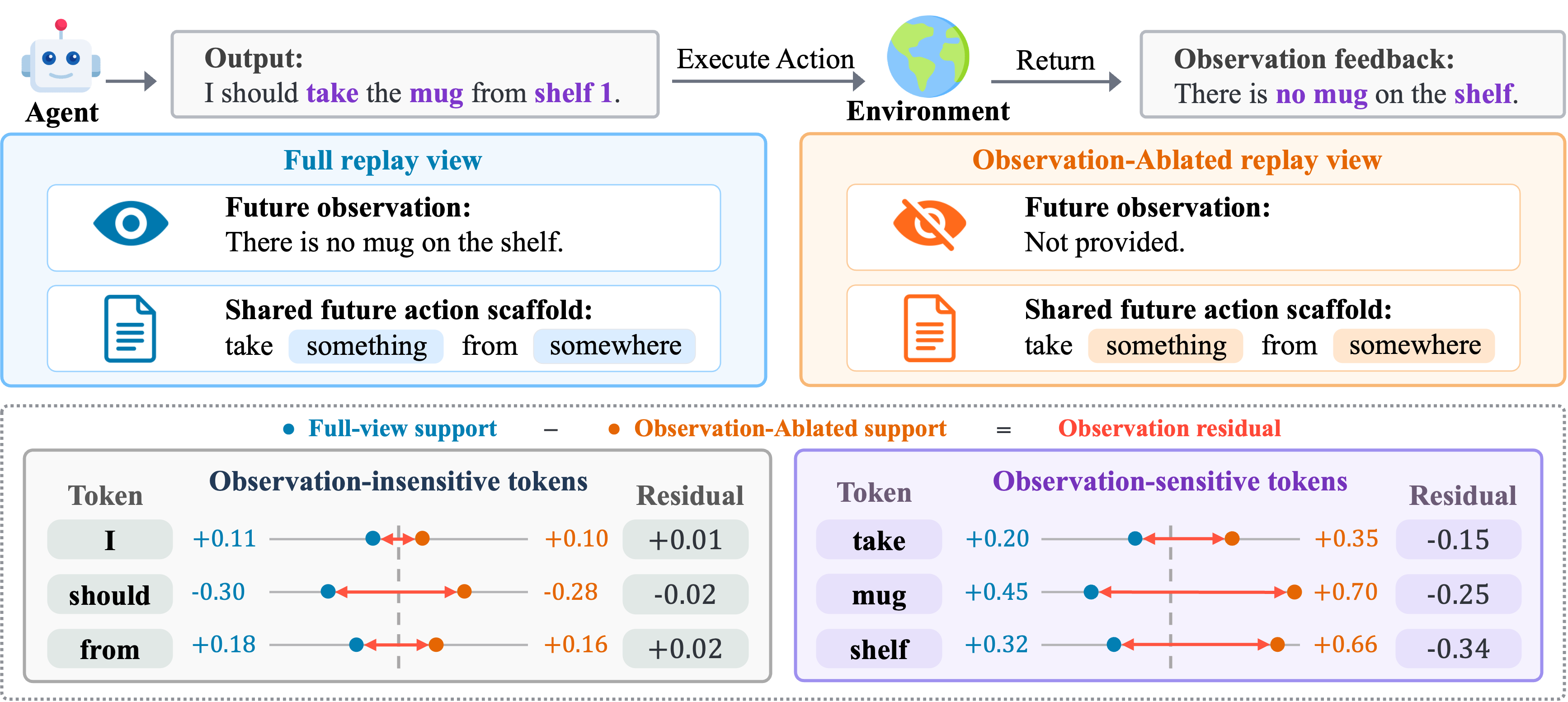}
    \caption{Deriving the observation residual from structurally matched replay views.}
    \label{fig:motivation}
    \vspace{-2mm}
\end{figure*}

This issue is particularly evident when future environment observations serve as privileged information. As shown in Figure~\ref{fig:motivation}, the agent incorrectly generates ``take the mug from shelf 1,'' after which the environment reports ``there is no mug on the shelf.'' The Full replay view incorporates this observation into the context and re-scores the original output. Still, in doing so, it also introduces an additional replay format and future-action scaffold. Even after removing the true observation, the Observation-Ablated view still assigns similar support to the Full view for some tokens, indicating that these changes are not entirely attributable to the environment feedback. Full-view support therefore conflates changes tied to the future observation with those shared by the replay scaffold. Controlling for the latter requires a structurally matched view that preserves the same replay structure while excluding the actual future observation.

% Accordingly, we propose \textbf{Observation-Calibrated Self-Distillation (OCSD)} for agentic reinforcement learning. OCSD constructs two structurally matched replay views, Full and Observation-Ablated, that differ only in whether the actual future observation is included. Comparing their scores for the same generated token yields an observation-sensitive residual after controlling for the shared replay scaffold. OCSD then applies this residual at high-uncertainty steps to modulate the token-level update strength of GRPO while preserving the trajectory-level update direction. In this way, future observations provide calibrated local supervision during training.

Accordingly, we propose \textbf{Observation-Calibrated Self-Distillation (OCSD)} for agentic reinforcement learning. OCSD constructs two structurally matched replay views, Full and Observation-Ablated, that differ only in whether the actual future observation is included, and contrasts their scores for the same generated token to derive an observation residual that discounts score changes shared by the replay scaffold. This residual is then applied at high-uncertainty steps to modulate the token-level update strength of GRPO, while the trajectory-level update direction is left unchanged. In this way, OCSD turns future observations into calibrated, step-local supervision rather than a confounded replay signal. Fine-grained diagnostic analyses further show that the calibrated residual aligns more closely with local environment feedback than Full-view support (Section~\ref{sec:token_diagnosis}).

% We evaluate OCSD on ALFWorld~\cite{alfworld}, WebShop~\cite{webshop}, and Search-QA~\cite{searchqa} using Qwen3 models at three different scales. OCSD achieves the best overall performance across all tasks and model scales, improving over GRPO by 8.9 to 14.0 percentage points on ALFWorld. Further analysis shows that the calibrated observation residual is more consistent with local environment feedback. Our contributions are threefold:
% (1) \textbf{Confounding in agentic replay scoring.}  We identify that Full-view support in agentic privileged replay scoring can reflect both future-observation content and changes shared by the replay scaffold.
% (2) \textbf{Observation-Calibrated Self-Distillation.} We propose OCSD, which derives an observation residual from structurally matched Full and Observation-Ablated views and applies it to token-level policy optimization at high-uncertainty steps.
% (3) \textbf{Systematic validation.} Results show that the residual better aligns with local environment feedback and that OCSD achieves consistent gains across three benchmarks and three model scales.

We evaluate OCSD on ALFWorld~\cite{alfworld}, WebShop~\cite{webshop}, and Search-QA~\cite{searchqa} using Qwen3 models at three different scales. OCSD achieves the best overall performance across all tasks and model scales. Our contributions are summarized as follows:
(1) We identify a confounding effect in agentic privileged replay scoring: Full-view support can reflect both information from future observations and score changes induced by the shared replay scaffold.
% (1) We identify that Full-view support in agentic privileged replay scoring can reflect both information from future observations and changes shared by the replay scaffold.
(2) We propose OCSD, which derives an observation residual from structurally matched Full and Observation-Ablated views and applies it to token-level policy optimization at high-uncertainty steps.
(3) Comprehensive experiments show that the residual better aligns with local environment feedback and that OCSD consistently outperforms strong baselines across three benchmarks and three model scales.

\section{Revisiting Privileged Replay Scoring}
\subsection{Preliminaries}
\paragraph{LLM Agents.}
Given a task $x$, an LLM agent interacts with the environment over a sequence
of steps. At step $k$, let $h_k$ denote the interaction context available to
the agent, including the task, interaction history, and current observation
$o_k$. The agent generates
$y_k=(y_{k,1},\ldots,y_{k,T_k})\sim\pi_\theta(\cdot\mid h_k)$, which
determines an executable action, after which the environment returns the next
observation $o_{k+1}$. We denote the context for generating token $y_{k,t}$
by $h_{k,t}=(h_k,y_{k,<t})$. The resulting trajectory $\tau$ receives a
terminal reward $R(\tau)$. When multiple trajectories are sampled, we add the
trajectory index $i$ to these quantities and let $K_i$ denote the number of
interaction steps in $\tau_i$.

% Given a task $x$, an LLM agent completes it through multiple rounds of
% interaction with the environment. At interaction step $k$, the agent receives
% an environment observation $o_k$. We denote the interaction context available
% at this step by
% \begin{equation}
% \label{eq:agent_context}
%     h_k
%     =
%     \left(
%         x,
%         o_1, y_1,
%         \ldots,
%         o_{k-1}, y_{k-1},
%         o_k
%     \right).
% \end{equation}
% Conditioned on $h_k$, the agent generates an output
% $y_k \sim \pi_\theta(\cdot \mid h_k)$, represented as a sequence of
% $T_k$ tokens:
% \begin{equation}
% \label{eq:step_output}
%     y_k
%     =
%     \left(
%         y_{k,1},
%         \ldots,
%         y_{k,T_k}
%     \right).
% \end{equation}
% We further denote the context for generating the $t$-th token by
% $h_{k,t}=(h_k,y_{k,<t})$.
% The output $y_k$ determines an executable action $a_k$. After the action is
% executed, the environment returns the next observation $o_{k+1}$. The complete
% interaction forms a trajectory
% \begin{equation}
% \label{eq:agent_trajectory}
%     \tau
%     =
%     \left(
%         x,
%         o_1, y_1, a_1,
%         \ldots,
%         o_K, y_K, a_K,
%         o_{K+1}
%     \right),
% \end{equation}
% which receives a terminal reward $R(\tau)$.

\paragraph{Group Relative Policy Optimization.}
For each task $x$, GRPO~\cite{grpo} samples a group of $G$ trajectories
$\{\tau_i\}_{i=1}^{G}$ from the rollout policy and computes a
group-relative advantage from their terminal rewards:
\begin{equation}
\label{eq:grpo_advantage}
\widehat{A}_i
=
\frac{
R(\tau_i)
-
\frac{1}{G}\sum_{j=1}^{G}R(\tau_j)
}{
\operatorname{std}
\left(
\{R(\tau_j)\}_{j=1}^{G}
\right)
+
\epsilon_{\mathrm{adv}}
}.
\end{equation}

GRPO assigns the same trajectory-level advantage $\widehat{A}_i$ to all
tokens $y_{i,k,t}$ in $\tau_i$, without distinguishing their individual
contributions to the terminal reward.

\begin{figure*}[t]
    \centering
    \includegraphics[width=\textwidth]{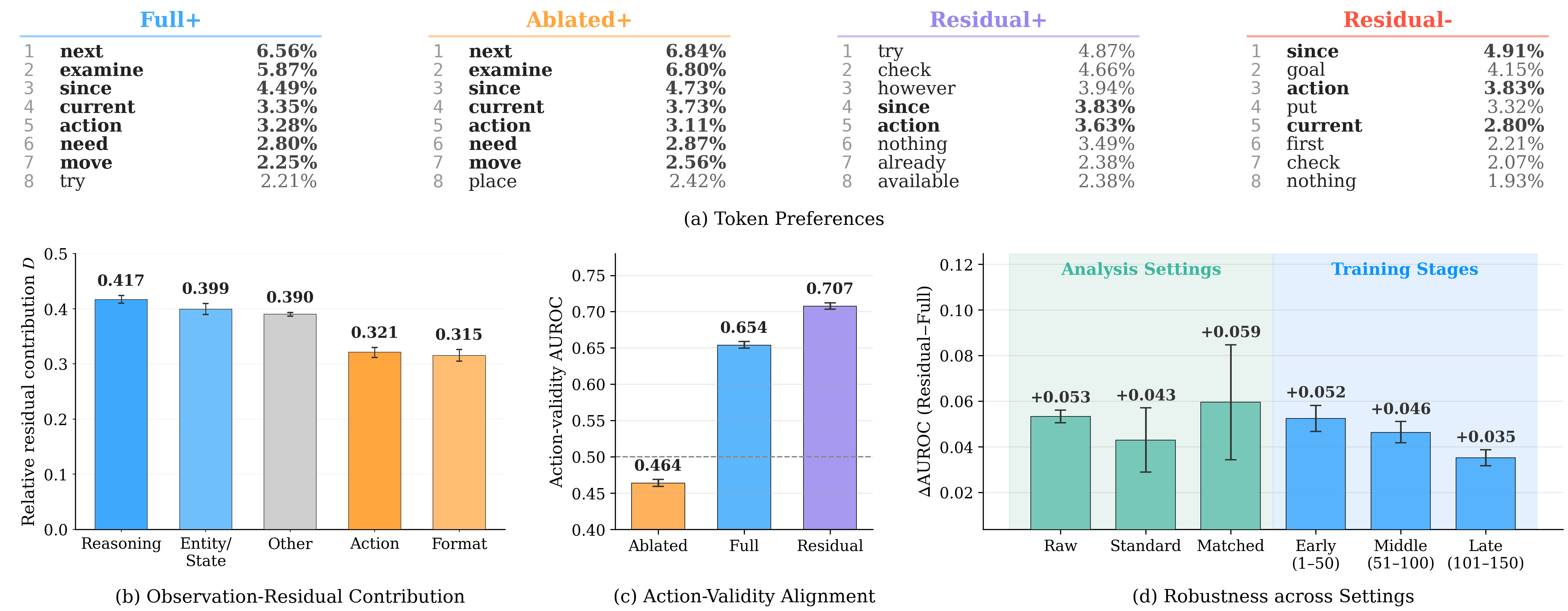}
    \caption{
    Fine-grained diagnosis of Qwen3-1.7B over 150 training steps on ALFWorld.
    All panels use the top-$20\%$ high-NLL interaction steps selected within
    each trajectory. In (b)--(d), error bars indicate trajectory-cluster
    bootstrap 95\% confidence intervals.
    (a) Words enriched in the upper $5\%$ tails of the checkpoint-centered
    Full, Observation-Ablated, and residual signals, and in the lower $5\%$
    tail of the residual.
    (b) Word-level relative residual contribution $D$ across five functional
    categories.
    (c) AUROC of step-averaged Observation-Ablated, Full, and residual signals
    for distinguishing valid from invalid actions; ambiguous cases are
    excluded.
    (d) AUROC improvement of the residual over Full support. Raw uses the
    original scores, Standard applies within-checkpoint normalization, and
    Matched controls for training stage and action operator. The final three
    bars report results at different training stages.
    }
    \label{fig:token_diagnosis}
    \vspace{-2mm}
\end{figure*}

\paragraph{Privileged Replay Scoring.}
OPSD generates on-policy outputs under the original interaction context and
re-scores the same tokens after rollout using replay evidence $E_{i,k}$.
We denote the original-context and replay-conditioned token distributions by
$\pi_S$ and $\pi_T$, referred to as the student and teacher views,
respectively. For token $y_{i,k,t}$ at the $k$-th interaction step of
trajectory $\tau_i$, its token-level support is defined as
\begin{equation}
\label{eq:privileged_token_support}
\begin{aligned}
\delta_{i,k,t}(E_{i,k})
={}&
\log \pi_T
\left(
    y_{i,k,t}
    \mid
    h_{i,k,t}, E_{i,k}
\right)
\\
&-
\log \pi_S
\left(
    y_{i,k,t}
    \mid
    h_{i,k,t}
\right).
\end{aligned}
\end{equation}

Positive values indicate increased support, while negative values indicate reduced support.

For agent tasks, we instantiate two structurally matched replay views. The
Full evidence $E^F_{i,k}$ contains the actual future observations together
with naturalized schemas of future actions. Each action schema preserves the
coarse action intent and syntactic structure while replacing high-information
argument values, such as concrete entities, locations, queries, and answers,
with natural-language expressions of their coarse semantic roles. The
Observation-Ablated evidence $E^A_{i,k}$ replaces only the
future-observation content with the fixed phrase ``Observation: not
provided,'' while keeping the field order, replay format, and future-action
scaffold unchanged. Their corresponding token-level supports are
\begin{equation}
\label{eq:full_ablated_support}
\delta^F_{i,k,t}
=
\delta_{i,k,t}(E^F_{i,k}),
\qquad
\delta^A_{i,k,t}
=
\delta_{i,k,t}(E^A_{i,k}).
\end{equation}
Under the Full and Observation-Ablated views, $\pi_T$ is instantiated as
$\pi_F$ and $\pi_A$, respectively, using the same model parameters with
different input contexts.

\subsection{Fine-Grained Diagnosis of Replay Support}
\label{sec:token_diagnosis}
As shown in Figure~\ref{fig:motivation}, the Full and
Observation-Ablated views assign nearly identical support to tokens such as
\texttt{I}, \texttt{should}, and \texttt{from}, but differ in their support for tokens such as \texttt{take}, \texttt{mug}, and \texttt{shelf}, which are more closely related to the actual future observation.

To examine whether the pattern in Figure~\ref{fig:motivation} generalizes
beyond the single example, we analyze the top-$20\%$ interaction steps ranked
by average token NLL within each trajectory, following the step-selection
procedure in Section~\ref{sec:uncertainty_step_selection}, over 150 training
steps of Qwen3-1.7B on ALFWorld. Within each checkpoint, we center
$\delta^F$, $\delta^A$, and $\delta^F-\delta^A$ by their respective means
and merge subword tokens into word occurrences by averaging their constituent
token scores. For each word, we compute the fraction of its occurrences that
fall in the upper $5\%$ tail of each centered signal, or in the lower $5\%$
tail for the negative residual, and aggregate these statistics equally across
checkpoints. Figure~\ref{fig:token_diagnosis}(a) scales each word by its
tail-hit rate multiplied by its log-scaled occurrence frequency. The upper tails of Full and Observation-Ablated support contain substantially overlapping words, whereas the observation residual highlights a distinct set. This suggests that Full-view support contains score changes shared by the replay scaffold, whereas the observation residual is more sensitive to the actual future observation.

To further examine how this difference varies across functional roles, for
each word occurrence $w$ with constituent token set
$\mathcal{T}_{i,k,w}$, we define
\begin{equation}
\label{eq:relative_residual_contribution}
\begin{aligned}
m^R_{i,k,w}
&=
\operatorname{mean}_{t\in\mathcal{T}_{i,k,w}}
\left|
\delta^F_{i,k,t}-\delta^A_{i,k,t}
\right|,
\\
m^A_{i,k,w}
&=
\operatorname{mean}_{t\in\mathcal{T}_{i,k,w}}
\left|
\delta^A_{i,k,t}
\right|,
\\
D_{i,k,w}
&=
\frac{
m^R_{i,k,w}
}{
m^R_{i,k,w}+m^A_{i,k,w}+\epsilon
}.
\end{aligned}
\end{equation}
A larger $D_{i,k,w}$ indicates a greater relative contribution of the observation residual. We group the analyzed words into five
mutually exclusive categories using deterministic lexical and
output-structure rules. As shown in Figure~\ref{fig:token_diagnosis}(b),
reasoning and environment entity/state words have higher $D$ than action and
format words, indicating that the difference between the two replay views
varies across functional roles.

\begin{figure*}[htbp]
    \centering
    \includegraphics[width=0.85\textwidth]{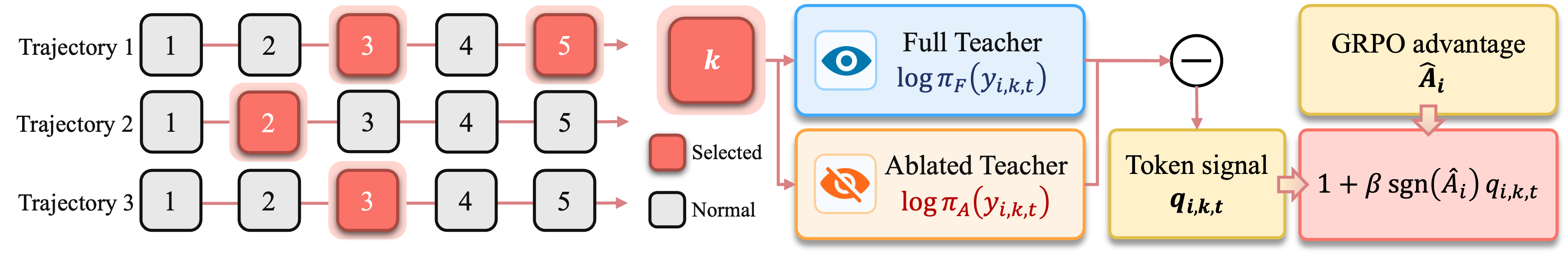}
    \caption{Overview of the OCSD training framework.}
    \label{fig:framework}
    \vspace{-2mm}
\end{figure*}

We next assess the consistency of the
three signals with local environment feedback. For each interaction step, we
average the signal over all generated tokens:
\begin{equation}
\label{eq:step_level_score}
S_{i,k}(s)
=
\frac{1}{T_{i,k}}
\sum_{t=1}^{T_{i,k}} s_{i,k,t},
\quad
s \in
\left\{
\delta^A,\,
\delta^F,\,
\delta^F-\delta^A
\right\}.
\end{equation}
Using deterministic rules based on the environment execution status and
templated feedback, we label actions that produce a state change as valid and
actions that are rejected or produce no effect as invalid, while excluding
ambiguous cases. We pool the labeled steps across checkpoints and evaluate
each signal using AUROC~\cite{auroc} with trajectory-cluster bootstrap. As
shown in Figure~\ref{fig:token_diagnosis}(c), the Observation-Ablated, Full,
and residual signals achieve AUROC values of $0.464$, $0.654$, and $0.707$,
respectively, showing that the residual is more consistent with local
environment feedback. Figure~\ref{fig:token_diagnosis}(d) further shows that
its improvement over Full support remains stable under within-checkpoint
standardization, stratified matching, and across training stages.

Together, these analyses show that Full support contains substantial
preferences that persist without the actual future observation, whereas the difference between the two views is more closely aligned with local
environment feedback.

\section{Methodology}
\subsection{Observation-Calibrated Token Signal}
\label{sec:observation_calibrated_signal}

Based on Full replay support $\delta^F_{i,k,t}$ and
Observation-Ablated support $\delta^A_{i,k,t}$, we define the
observation residual as:
\begin{equation}
\label{eq:observation_residual}
e_{i,k,t}
=
\delta^F_{i,k,t}
-
\delta^A_{i,k,t}.
\end{equation}

Both support terms are computed relative to the same student prediction, so
the student log-probability cancels in their difference. Expanding the two
terms gives
\begin{equation}
\label{eq:observation_residual_expanded}
\begin{aligned}
e_{i,k,t}
={}&
\log \pi_F
\bigl(
    y_{i,k,t}
    \mid
    h_{i,k,t}, E^F_{i,k}
\bigr)
\\
&-
\log \pi_A
\bigl(
    y_{i,k,t}
    \mid
    h_{i,k,t}, E^A_{i,k}
\bigr).
\end{aligned}
\end{equation}
Thus, $e_{i,k,t}$ directly compares the support assigned to the same token by
the Full Teacher and the Observation-Ablated Teacher, and captures the
incremental difference between the two replay views.

We further map the unbounded log-probability difference to a bounded
observation-calibrated token signal:
\begin{equation}
\label{eq:bounded_observation_signal}
q_{i,k,t}
=
\tanh
\left(
    \frac{e_{i,k,t}}{2}
\right),
\qquad
q_{i,k,t}\in[-1,1].
\end{equation}
The sign of $q_{i,k,t}$ indicates whether the Full Teacher assigns higher or
lower support to the current token than the Observation-Ablated Teacher, while
its magnitude reflects the degree of disagreement between the two replay
views.

\subsection{NLL-Guided Step Selection}
\label{sec:uncertainty_step_selection}

Following prior work on selective training~\cite{nll_1,nll_2}, we prioritize interaction steps with high average token NLL under the rollout policy.
For the $k$-th interaction step in trajectory $\tau_i$, we define the average
negative log-likelihood of all generated tokens under the old policy as the
step uncertainty score:
\begin{equation}
\label{eq:step_uncertainty}
u_{i,k}
=
-
\frac{1}{T_{i,k}}
\sum_{t=1}^{T_{i,k}}
\log
\pi_{\theta_{\mathrm{old}}}
\left(
    y_{i,k,t}
    \mid
    h_{i,k,t}
\right).
\end{equation}

A larger $u_{i,k}$ indicates greater uncertainty of the rollout policy about
the generated output at that step. Within each trajectory, we rank all
interaction steps in descending order of $u_{i,k}$ and select the top
$\rho$ fraction:
\begin{equation}
\label{eq:top_step_selection}
\mathcal{S}_i
=
\operatorname{TopIndices}_{\lceil \rho K_i\rceil}
\left(
    \left\{
        u_{i,k}
    \right\}_{k=1}^{K_i}
\right),
\end{equation}
where $\rho\in(0,1]$ is the step selection ratio. The ceiling operation ensures
that at least one step is selected from each trajectory. OCSD computes the
observation-calibrated signal for all generated tokens within each selected
step.

\subsection{Sign-Preserving Integration with GRPO}
\label{sec:sign_preserving_grpo}

OCSD preserves the update direction determined by the trajectory-level GRPO
advantage, while using the observation-calibrated signal to adjust the update
strength of individual tokens within the selected steps. For token
$y_{i,k,t}$ in trajectory $\tau_i$, we define the calibrated advantage as
\begin{equation}
\label{eq:ocsd_advantage}
\widehat{A}^{\mathrm{OCSD}}_{i,k,t}
=
\begin{cases}
\widehat{A}_i
\left[
    1
    +
    \beta
    \operatorname{sgn}(\widehat{A}_i)
    q_{i,k,t}
\right],
&
k\in\mathcal{S}_i,
\\[0.4em]
\widehat{A}_i,
&
k\notin\mathcal{S}_i,
\end{cases}
\end{equation}
where $\beta\in[0,1]$ controls the strength of observation calibration.
OCSD modulates all generated tokens within each selected step, while tokens
in unselected steps retain the original GRPO advantage. Since
$q_{i,k,t}\in[-1,1]$, the modulation factor is non-negative, so
$\widehat{A}^{\mathrm{OCSD}}_{i,k,t}$ preserves the sign of
$\widehat{A}_i$.

For token $y_{i,k,t}$, the importance ratio is defined as
\begin{equation}
\label{eq:ocsd_importance_ratio}
r_{i,k,t}(\theta)
=
\frac{
\pi_\theta
\left(
    y_{i,k,t}
    \mid
    h_{i,k,t}
\right)
}{
\pi_{\theta_{\mathrm{old}}}
\left(
    y_{i,k,t}
    \mid
    h_{i,k,t}
\right)
}.
\end{equation}

We then replace the trajectory-level advantage in the standard GRPO surrogate
with the calibrated token-level advantage:
\begin{equation}
\label{eq:ocsd_objective}
\begin{aligned}
\mathcal{J}^{\mathrm{OCSD}}(\theta)
={}&
\mathbb{E}_{i,k,t}
\Bigg[
\min\Big(
r_{i,k,t}(\theta)
\widehat{A}^{\mathrm{OCSD}}_{i,k,t},
\\[-0.2em]
&
% \hspace{1.0em}
\operatorname{clip}\!\left(
    r_{i,k,t}(\theta),
    1-\epsilon,
    1+\epsilon
\right)
\widehat{A}^{\mathrm{OCSD}}_{i,k,t}
\Big)
\\[-0.2em]
&
% \hspace{1.0em}
-
\beta_{\mathrm{KL}}
\mathrm{D}_{\mathrm{KL}}\!\left(
    \pi_\theta
    \Vert
    \pi_{\mathrm{ref}}
\right)_{i,k,t}
\Bigg].
\end{aligned}
\end{equation}
Here, $\pi_{\mathrm{ref}}$ is the reference policy, while $\epsilon$ and
$\beta_{\mathrm{KL}}$ control the clipping range and KL regularization,
respectively. The expectation is taken over all sampled trajectories and valid
generated tokens.

\begin{table*}[!t]
\centering

{
\small
\setlength{\tabcolsep}{1pt}

\begin{tabular*}{\textwidth}{
@{\extracolsep{\fill}}
l
l
ccccccc
@{\hspace{4pt}}
cc
@{}
}
\toprule
\multirow{2}{*}{\textbf{Model}} &
\multirow{2}{*}{\textbf{Method}} &
\multicolumn{7}{c}{\textbf{ALFWorld}} &
\multicolumn{2}{c}{\textbf{WebShop}} \\
\cmidrule(lr){3-9}
\cmidrule(lr){10-11}
&
&
\textbf{Pick} &
\textbf{Look} &
\textbf{Clean} &
\textbf{Heat} &
\textbf{Cool} &
\textbf{Pick2} &
\textbf{Overall} &
\textbf{Score} &
\textbf{Success} \\
\midrule

\multirow{7}{*}{\textit{Qwen3-1.7B}}
& Vanilla
& 13.8 & 33.3 & 3.2 & 10.5 & 0.0 & 5.0 & 7.8
& 44.0 & 2.3 \\

& GRPO
& {\cellcolor{second}\meanstd{70.4}{8.0}}
& \meanstd{44.0}{1.7}
& \meanstd{40.6}{7.4}
& \meanstd{38.6}{5.8}
& \meanstd{33.5}{5.8}
& \meanstd{30.1}{7.3}
& \meanstd{46.6}{0.7}
& \meanstd{71.8}{1.0}
& \meanstd{48.4}{0.8} \\

& OPSD
& \meanstd{33.9}{3.5}
& \meanstd{30.7}{7.9}
& \meanstd{5.4}{1.8}
& \meanstd{1.6}{2.3}
& \meanstd{2.9}{2.0}
& \meanstd{11.6}{5.0}
& \meanstd{14.6}{0.3}
& \meanstd{50.3}{0.6}
& \meanstd{11.4}{0.5} \\

& GRPO+OPSD
& \meanstd{63.1}{7.2}
& \meanstd{41.5}{5.8}
& \meanstd{39.2}{9.2}
& \meanstd{44.8}{9.4}
& \meanstd{27.3}{7.5}
& \meanstd{29.1}{1.8}
& \meanstd{41.7}{2.4}
& \meanstd{72.1}{2.6}
& \meanstd{49.0}{1.9} \\

& RLSD
& \meanstd{49.9}{6.2}
& {\cellcolor{second}\meanstd{51.9}{2.6}}
& \meanstd{46.7}{3.6}
& {\cellcolor{second}\meanstd{47.5}{7.4}}
& \meanstd{38.1}{3.0}
& \meanstd{26.9}{4.7}
& \meanstd{44.5}{1.7}
& {\cellcolor{best}\bfseries\boldmath\meanstd{76.3}{2.5}}
& \meanstd{52.1}{2.9} \\

& SDAR
& \meanstd{66.7}{9.7}
& \meanstd{50.6}{7.5}
& {\cellcolor{second}\meanstd{51.4}{7.9}}
& \meanstd{33.7}{6.5}
& {\cellcolor{second}\meanstd{43.1}{3.0}}
& {\cellcolor{second}\meanstd{38.1}{3.6}}
& {\cellcolor{second}\meanstd{51.6}{2.2}}
& \meanstd{73.4}{1.0}
& {\cellcolor{second}\meanstd{53.4}{1.0}} \\

& \textbf{OCSD (ours)}
& {\cellcolor{best}\bfseries\boldmath\meanstd{75.9}{4.9}}
& {\cellcolor{best}\bfseries\boldmath\meanstd{63.1}{2.7}}
& {\cellcolor{best}\bfseries\boldmath\meanstd{51.9}{4.7}}
& {\cellcolor{best}\bfseries\boldmath\meanstd{47.7}{4.3}}
& {\cellcolor{best}\bfseries\boldmath\meanstd{46.4}{1.8}}
& {\cellcolor{best}\bfseries\boldmath\meanstd{42.3}{6.9}}
& {\cellcolor{best}\bfseries\boldmath\meanstd{55.5}{2.3}}
& {\cellcolor{second}\meanstd{74.7}{3.4}}
& {\cellcolor{best}\bfseries\boldmath\meanstd{54.4}{1.2}} \\

\midrule

\multirow{7}{*}{\textit{Qwen3-4B}}
& Vanilla
& 60.0 & 33.3 & 16.7 & 31.6 & 0.0 & 20.0 & 27.3
& 16.2 & 3.9 \\

& GRPO
& \meanstd{86.6}{0.6}
& \meanstd{59.5}{5.1}
& \meanstd{77.7}{4.7}
& \meanstd{67.1}{4.1}
& \meanstd{55.5}{4.1}
& \meanstd{55.0}{6.2}
& \meanstd{70.6}{1.3}
& \meanstd{81.9}{3.6}
& \meanstd{69.5}{1.5} \\

& OPSD
& \meanstd{69.9}{3.1}
& \meanstd{39.2}{4.5}
& \meanstd{26.6}{4.8}
& \meanstd{26.6}{4.3}
& \meanstd{19.0}{4.6}
& \meanstd{26.4}{3.8}
& \meanstd{36.2}{1.9}
& \meanstd{27.8}{2.9}
& \meanstd{4.7}{0.6} \\

& GRPO+OPSD
& {\cellcolor{second}\meanstd{88.0}{0.4}}
& \meanstd{52.8}{3.9}
& \meanstd{75.0}{4.1}
& \meanstd{62.5}{3.4}
& \meanstd{68.3}{8.1}
& \meanstd{36.9}{2.5}
& \meanstd{68.8}{2.6}
& \meanstd{82.5}{1.0}
& \meanstd{71.3}{1.2} \\

& RLSD
& \meanstd{87.9}{4.0}
& {\cellcolor{best}\bfseries\boldmath\meanstd{83.4}{1.8}}
& {\cellcolor{second}\meanstd{80.5}{0.9}}
& \meanstd{70.3}{9.4}
& {\cellcolor{second}\meanstd{79.6}{4.0}}
& {\cellcolor{second}\meanstd{60.6}{4.3}}
& {\cellcolor{second}\meanstd{79.2}{1.0}}
& {\cellcolor{second}\meanstd{83.7}{1.5}}
& \meanstd{68.5}{1.2} \\

& SDAR
& \meanstd{85.2}{4.8}
& \meanstd{63.9}{4.0}
& \meanstd{76.8}{5.8}
& {\cellcolor{second}\meanstd{80.0}{3.6}}
& \meanstd{72.9}{4.4}
& \meanstd{42.6}{6.9}
& \meanstd{73.2}{1.3}
& \meanstd{83.6}{1.8}
& {\cellcolor{second}\meanstd{71.9}{0.7}} \\

& \textbf{OCSD (ours)}
& {\cellcolor{best}\bfseries\boldmath\meanstd{91.9}{1.4}}
& {\cellcolor{second}\meanstd{77.2}{4.4}}
& {\cellcolor{best}\bfseries\boldmath\meanstd{90.2}{4.6}}
& {\cellcolor{best}\bfseries\boldmath\meanstd{84.7}{4.4}}
& {\cellcolor{best}\bfseries\boldmath\meanstd{80.9}{4.3}}
& {\cellcolor{best}\bfseries\boldmath\meanstd{63.8}{0.9}}
& {\cellcolor{best}\bfseries\boldmath\meanstd{82.8}{0.7}}
& {\cellcolor{best}\bfseries\boldmath\meanstd{86.9}{1.7}}
& {\cellcolor{best}\bfseries\boldmath\meanstd{73.7}{1.6}} \\

\midrule

\multirow{7}{*}{\textit{Qwen3-8B}}
& Vanilla
& 58.6 & 50.0 & 22.6 & 21.1 & 34.8 & 45.0 & 37.5
& 17.9 & 4.7 \\

& GRPO
& \meanstd{86.8}{4.5}
& \meanstd{44.4}{7.9}
& \meanstd{81.5}{6.9}
& \meanstd{67.8}{0.8}
& \meanstd{70.4}{8.5}
& \meanstd{63.2}{3.4}
& \meanstd{73.2}{4.0}
& \meanstd{85.4}{3.1}
& \meanstd{75.5}{1.2} \\

& OPSD
& \meanstd{62.3}{3.7}
& \meanstd{44.8}{4.5}
& \meanstd{54.2}{4.2}
& \meanstd{18.7}{5.6}
& \meanstd{38.8}{2.9}
& \meanstd{38.4}{0.8}
& \meanstd{48.2}{1.2}
& \meanstd{17.1}{0.7}
& \meanstd{3.9}{0.0} \\

& GRPO+OPSD
& \meanstd{84.7}{4.3}
& \meanstd{63.9}{3.9}
& \meanstd{83.9}{2.4}
& \meanstd{65.5}{2.9}
& \meanstd{85.1}{6.0}
& \meanstd{70.7}{4.2}
& \meanstd{78.6}{1.4}
& \meanstd{85.3}{2.1}
& {\cellcolor{second}\meanstd{76.1}{0.9}} \\

& RLSD
& {\cellcolor{second}\meanstd{90.4}{2.9}}
& {\cellcolor{second}\meanstd{77.5}{4.9}}
& {\cellcolor{best}\bfseries\boldmath\meanstd{94.1}{2.1}}
& \meanstd{80.5}{3.8}
& {\cellcolor{second}\meanstd{85.3}{2.0}}
& {\cellcolor{second}\meanstd{73.6}{2.0}}
& {\cellcolor{second}\meanstd{85.4}{1.0}}
& \meanstd{85.4}{0.9}
& \meanstd{74.2}{0.7} \\

& SDAR
& \meanstd{87.5}{2.8}
& \meanstd{58.2}{3.6}
& \meanstd{81.0}{4.3}
& {\cellcolor{second}\meanstd{80.7}{6.8}}
& \meanstd{64.2}{5.0}
& \meanstd{60.3}{2.8}
& \meanstd{75.0}{1.1}
& {\cellcolor{second}\meanstd{86.6}{1.0}}
& \meanstd{75.0}{0.8} \\

& \textbf{OCSD (ours)}
& {\cellcolor{best}\bfseries\boldmath\meanstd{92.8}{2.2}}
& {\cellcolor{best}\bfseries\boldmath\meanstd{92.6}{6.9}}
& {\cellcolor{second}\meanstd{90.4}{7.0}}
& {\cellcolor{best}\bfseries\boldmath\meanstd{87.8}{8.8}}
& {\cellcolor{best}\bfseries\boldmath\meanstd{85.5}{4.0}}
& {\cellcolor{best}\bfseries\boldmath\meanstd{74.2}{5.8}}
& {\cellcolor{best}\bfseries\boldmath\meanstd{87.2}{1.0}}
& {\cellcolor{best}\bfseries\boldmath\meanstd{87.4}{1.9}}
& {\cellcolor{best}\bfseries\boldmath\meanstd{78.1}{0.8}} \\

\bottomrule
\end{tabular*}
}

\caption{Performance on ALFWorld and WebShop tasks. Results are reported as mean $\pm$ standard deviation over three runs with different random seeds for most experiments. Blue and red backgrounds denote the best and second-best methods, respectively.}
\label{tab:main_results}
\vspace{-2mm}
\end{table*}

\section{Experiments}

% \midrule
% \rowcolor[gray]{.95} \multicolumn{10}{l}{\textit{R1-Distill-7B}} \\
% Vanilla & 0.0 & 0.0 & 0.0 & 0.0 & 0.0 & 0.0 & 0.0 & 47.2 & 6.2 \\
% OPSD & \meanstd{0}{0} & \meanstd{0}{0} & \meanstd{0}{0} & \meanstd{0}{0} & \meanstd{0}{0} & \meanstd{0}{0} & \meanstd{0}{0} & \meanstd{45.1}{0.5} & \meanstd{8.1}{0.4}  \\
% GRPO & \meanstd{0}{0} & \meanstd{0}{0} & \meanstd{0}{0} & \meanstd{0}{0} & \meanstd{0}{0} & \meanstd{0}{0} & \meanstd{0}{0} & \meanstd{80.3}{1.6} & \meanstd{64.8}{0.6} \\
% GRPO+OPSD & \meanstd{0}{0} & \meanstd{0}{0} & \meanstd{0}{0} & \meanstd{0}{0} & \meanstd{0}{0} & \meanstd{0}{0} & \meanstd{0}{0} & \meanstd{0}{0} & \meanstd{0}{0} \\
% SDAR & \meanstd{0}{0} & \meanstd{0}{0} & \meanstd{0}{0} & \meanstd{0}{0} & \meanstd{0}{0} & \meanstd{0}{0} & \meanstd{0}{0} &  \meanstd{81.8}{1.2} &\meanstd{69.3}{1.3} \\
% \textbf{OCSD(ours)} & \meanstd{0}{0} & \meanstd{0}{0} & \meanstd{0}{0} & \meanstd{0}{0} & \meanstd{0}{0} & \meanstd{0}{0} & \meanstd{0}{0} & \meanstd{86.5}{1.1} & \meanstd{71.1}{1.1} \\

\subsection{Experimental Setup}
\paragraph{Benchmarks.}
We conduct experiments on ALFWorld~\cite{alfworld}, WebShop~\cite{webshop}, and Search-QA~\cite{searchqa}.
ALFWorld is a text-based embodied benchmark with six long-horizon household
manipulation tasks. WebShop evaluates product search, filtering, and purchase
under user-specified requirements. SearchQA evaluates external-information
retrieval and integration across three factoid and four multi-hop
question-answering datasets.

\paragraph{Baselines.}
We compare three categories of methods:
(1) \textbf{base models and reward-only reinforcement learning},
including the Vanilla model without task-specific post-training and
GRPO~\cite{grpo}, which is trained solely with environment rewards;
(2) \textbf{direct self-distillation baselines}, including
OPSD~\cite{opsd}, which performs self-distillation without reinforcement
learning, and GRPO+OPSD, which directly combines the two objectives;
and (3) \textbf{dedicated RL--OPSD integration methods}, including
RLSD~\cite{rlsd}, which uses self-distillation signals to modulate RL
updates, and SDAR~\cite{sdar}, which introduces a gated auxiliary
distillation objective.

\paragraph{Implementation Details.}
We conduct experiments across three model scales in the Qwen3
family~\cite{qwen3} and, following SDAR~\cite{sdar}, use a unified training
protocol for all post-training methods. For ALFWorld and WebShop, each batch
contains 16 tasks with 8 rollouts per task; for Search-QA, each batch contains
128 tasks. In Search-QA, NQ and HotpotQA serve as the in-domain training
distributions, while the remaining five datasets are used for out-of-domain
evaluation. We report Exact Match (EM), micro-averaged over the in-domain
(ID), out-of-domain (OOD), and all datasets (Avg.). For OCSD, we set the step
selection ratio to $\rho=0.2$ and the advantage modulation coefficient to
$\beta=0.5$. Additional training hyperparameters, evaluation details, and
hyperparameter sensitivity analyses are provided in the supplementary
material.

\subsection{Main Results}
\label{sec:main_results}
Tables~\ref{tab:main_results} and~\ref{tab:searchqa} report the main
results on ALFWorld, WebShop, and SearchQA. We draw the following
conclusions:

\paragraph{OCSD achieves the strongest overall results across all benchmarks.}
On ALFWorld, OCSD achieves overall success rates of 55.5, 82.8, and 87.2
with Qwen3-1.7B, 4B, and 8B, outperforming GRPO by 8.9, 12.2, and 14.0
points, respectively. It also achieves WebShop success rates of 54.4, 73.7, and 78.1, and SearchQA average EM scores of 43.1, 47.5, and 49.1.

\paragraph{Direct use of privileged replay support yields inconsistent gains.}
Directly using privileged replay support does not produce consistent improvements. Standalone OPSD performs substantially worse than GRPO across different models and environments, and exhibits particularly poor and unstable performance on SearchQA. Directly incorporating OPSD into GRPO also fails to guarantee stable gains. Although GRPO+OPSD improves performance in several settings, it still underperforms the original GRPO on ALFWorld with Qwen3-1.7B and Qwen3-4B.

\paragraph{OCSD provides broad improvements across ALFWorld task categories.}
Across the three model scales, OCSD achieves the best or second-best performance on most ALFWorld task categories, including basic object manipulation tasks such as Pick and more challenging tasks such as Pick2 that involve multi-step interactions.

\paragraph{OCSD performs consistently well under distribution shift.}
Across all three model scales, OCSD achieves the best OOD results among
the compared methods, suggesting that its gains are not confined to the
training distributions and extend to unseen retrieval and question-answering
datasets.

\begin{table}[!t]
\centering
\begin{tabular}{l ccc}
\toprule
\multirow{2}{*}{\textbf{Method}} &
\multicolumn{3}{c}{\textbf{SearchQA}} \\
\cmidrule(lr){2-4}
&
\textbf{ID} &
\textbf{OOD} &
\textbf{Avg.} \\
\midrule

\rowcolor[gray]{.95}
\multicolumn{4}{l}{\textit{Qwen3-1.7B}} \\

Vanilla
& 19.6
& 23.3
& 22.5 \\

GRPO
& \meanstd{36.3}{1.6}
& \meanstd{42.2}{0.6}
& \meanstd{40.9}{0.7} \\

OPSD
& \meanstd{1.4}{1.7}
& \meanstd{2.5}{2.9}
& \meanstd{2.3}{2.7} \\

GRPO+OPSD
& {\cellcolor{second}\meanstd{37.5}{1.7}}
& {\cellcolor{second}\meanstd{44.0}{0.2}}
& {\cellcolor{second}\meanstd{42.6}{0.4}} \\

RLSD
& \meanstd{36.7}{0.5}
& \meanstd{42.2}{0.8}
& \meanstd{41.0}{0.7} \\

SDAR
& \meanstd{36.9}{0.6}
& \meanstd{43.2}{0.9}
& \meanstd{41.8}{0.8} \\

\textbf{OCSD (ours)}
& {\cellcolor{best}\bfseries\boldmath\meanstd{37.6}{0.8}}
& {\cellcolor{best}\bfseries\boldmath\meanstd{44.5}{1.3}}
& {\cellcolor{best}\bfseries\boldmath\meanstd{43.1}{1.2}} \\

\midrule

\rowcolor[gray]{.95}
\multicolumn{4}{l}{\textit{Qwen3-4B}} \\

Vanilla
& 27.3
& 34.5
& 33.0 \\

GRPO
& \meanstd{42.1}{0.2}
& \meanstd{46.4}{0.9}
& \meanstd{45.5}{0.8} \\

OPSD
& \meanstd{0.3}{0.4}
& \meanstd{0.7}{1.0}
& \meanstd{0.6}{0.8} \\

GRPO+OPSD
& {\cellcolor{second}\meanstd{42.5}{1.1}}
& \meanstd{46.3}{1.3}
& \meanstd{45.5}{1.2} \\

RLSD
& {\cellcolor{best}\bfseries\boldmath\meanstd{42.9}{0.3}}
& {\cellcolor{second}\meanstd{47.2}{0.6}}
& {\cellcolor{second}\meanstd{46.3}{0.5}} \\

SDAR
& \meanstd{42.3}{0.2}
& \meanstd{46.9}{0.2}
& \meanstd{45.9}{0.2} \\

\textbf{OCSD (ours)}
& {\cellcolor{second}\meanstd{42.5}{0.9}}
& {\cellcolor{best}\bfseries\boldmath\meanstd{48.9}{1.0}}
& {\cellcolor{best}\bfseries\boldmath\meanstd{47.5}{0.8}} \\

\midrule

\rowcolor[gray]{.95}
\multicolumn{4}{l}{\textit{Qwen3-8B}} \\

Vanilla
& 29.1
& 37.4
& 35.6 \\

GRPO
& \meanstd{43.7}{0.1}
& {\cellcolor{second}\meanstd{50.1}{0.3}}
& {\cellcolor{second}\meanstd{48.7}{0.3}} \\

OPSD
& \meanstd{33.7}{0.8}
& \meanstd{41.1}{0.7}
& \meanstd{39.5}{0.6} \\

GRPO+OPSD
& {\cellcolor{second}\meanstd{44.2}{1.7}}
& \meanstd{49.3}{0.4}
& \meanstd{48.2}{0.2} \\

RLSD
& {\cellcolor{best}\bfseries\boldmath\meanstd{45.6}{0.5}}
& \meanstd{49.1}{0.1}
& \meanstd{48.4}{0.1} \\

SDAR
& \meanstd{43.8}{0.4}
& \meanstd{47.9}{0.3}
& \meanstd{47.0}{0.4} \\

\textbf{OCSD (ours)}
& \meanstd{43.6}{0.2}
& {\cellcolor{best}\bfseries\boldmath\meanstd{50.6}{0.4}}
& {\cellcolor{best}\bfseries\boldmath\meanstd{49.1}{0.3}} \\

\bottomrule
\end{tabular}
\caption{Performance on the SearchQA task.}
\vspace{-2mm}
\label{tab:searchqa}
\end{table}

\subsection{Ablation Analysis}

\begin{table}[!t]
\centering
\small
\setlength{\tabcolsep}{5pt}

\begin{tabular}{l c cc}
\toprule
\multirow{2}{*}{\textbf{Method}} &
\multicolumn{1}{c}{\textbf{ALFWorld}} &
\multicolumn{2}{c}{\textbf{WebShop}} \\
\cmidrule(lr){2-2}
\cmidrule(lr){3-4}
& \textbf{Overall} & \textbf{Score} & \textbf{Success} \\
\midrule

GRPO
& \meanstd{70.6}{1.3}
& \meanstd{81.9}{3.6}
& \meanstd{69.5}{1.5} \\

w/o Ablated Teacher
& \meanstd{74.8}{1.6}
& \meanstd{85.0}{1.0}
& \meanstd{72.9}{1.8} \\

w/ Random Select
& \meanstd{78.9}{1.1}
& \meanstd{84.2}{1.6}
& \meanstd{70.6}{2.6} \\

w/o Step Select
& \meanstd{66.4}{1.1}
& \meanstd{82.4}{2.5}
& \meanstd{69.8}{1.8} \\

w/o Sign Alignment
& \meanstd{71.4}{0.5}
& \meanstd{81.5}{2.4}
& \meanstd{71.6}{1.5} \\

\textbf{OCSD (ours)}
& \bfseries\boldmath\meanstd{82.8}{0.7}
& \bfseries\boldmath\meanstd{86.9}{1.7}
& \bfseries\boldmath\meanstd{73.7}{1.6} \\

\bottomrule
\end{tabular}

\caption{Qwen3-4B ablations on ALFWorld and WebShop.}
\label{tab:ablation}
\vspace{-2mm}
\end{table}

We evaluate four OCSD variants: (1) \textit{w/o Ablated Teacher} removes the
Observation-Ablated Teacher and directly uses the token-level signal from the
Full Teacher. (2) \textit{Random Selection} replaces high-NLL step selection with
random selection. (3) \textit{w/o Step Selection} applies residual calibration to
all interaction steps. (4) \textit{w/o Sign Alignment} removes the
advantage-dependent sign term in Eq.~\ref{eq:ocsd_advantage}, so the same
residual affects positive- and negative-advantage trajectories identically.

As shown in Table~\ref{tab:ablation}, all ablated variants underperform the complete OCSD method. Specifically, \textit{w/o Ablated Teacher} outperforms GRPO on both tasks, showing that future observations themselves provide useful supervision. \textit{Random Selection} also performs better than GRPO, indicating that residual calibration remains beneficial even when applied to randomly selected steps. In contrast, \textit{w/o Step Selection} falls below GRPO on ALFWorld and remains close to GRPO on WebShop, suggesting that indiscriminately calibrating all steps introduces low-value or noisy signals. Removing sign alignment similarly reduces ALFWorld performance and WebShop Score to around the GRPO level, confirming that the residual should be interpreted jointly with the direction of the trajectory-level advantage.

\begin{figure}[t]
    \centering
    \includegraphics[width=\columnwidth]{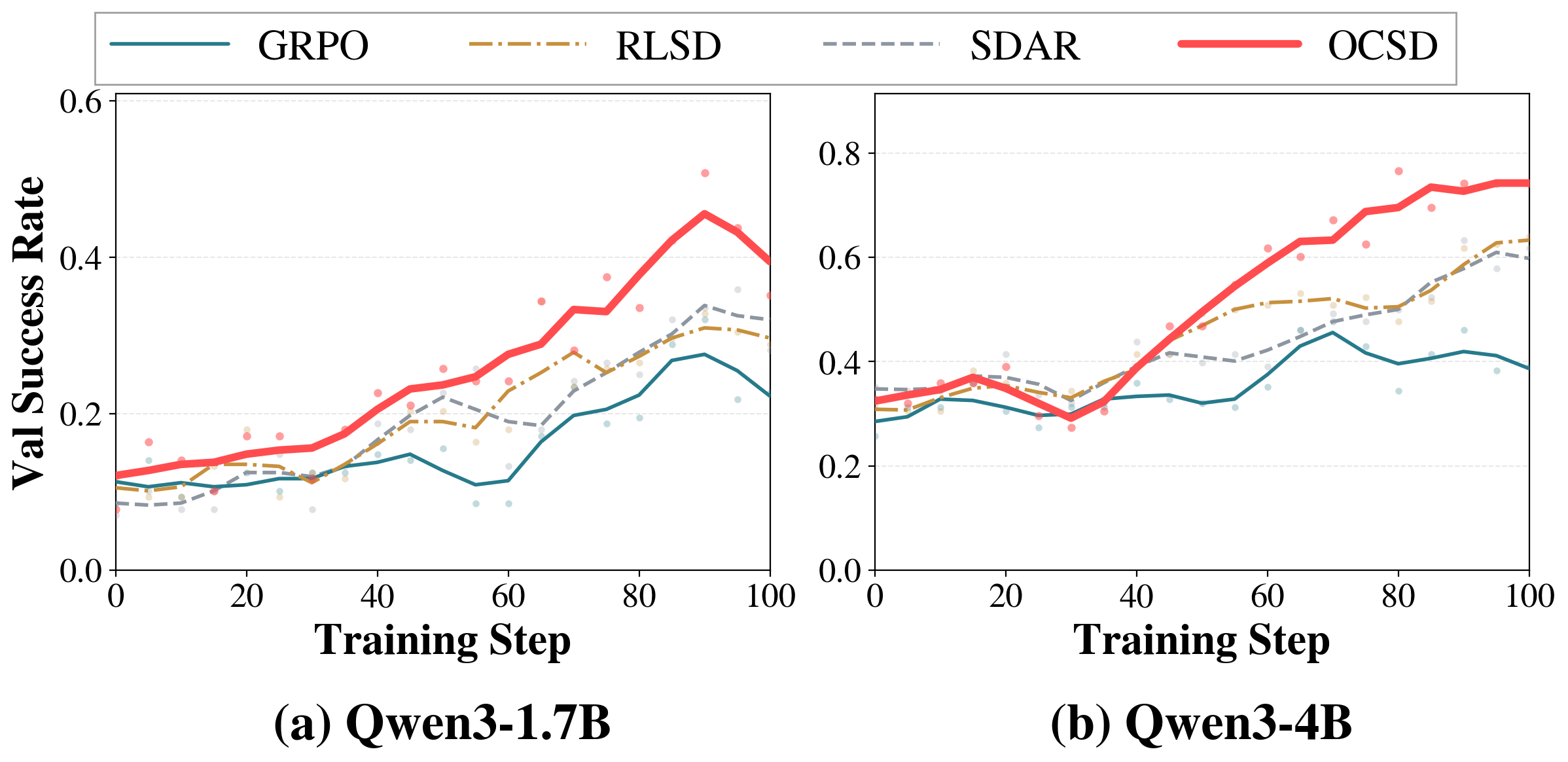}
    \caption{ALFWorld training dynamics across model scales.}
    \label{fig:training_dynamics}
    \vspace{-2mm}
\end{figure}

\subsection{More Analysis}
\paragraph{Comparison of Training Dynamics.}
Figure~\ref{fig:training_dynamics} compares the validation success rates of OCSD, GRPO, RLSD, and SDAR throughout training. Overall, OCSD achieves competitive or better performance over most of the displayed training period, with its advantage becoming more evident as training progresses.
During the early stage, the different methods perform similarly and
exhibit some fluctuations across adjacent checkpoints. As training
continues, OCSD improves more steadily and gradually separates from the
competing methods across both model scales.

\begin{figure}[t]
    \centering
    \includegraphics[width=\columnwidth]{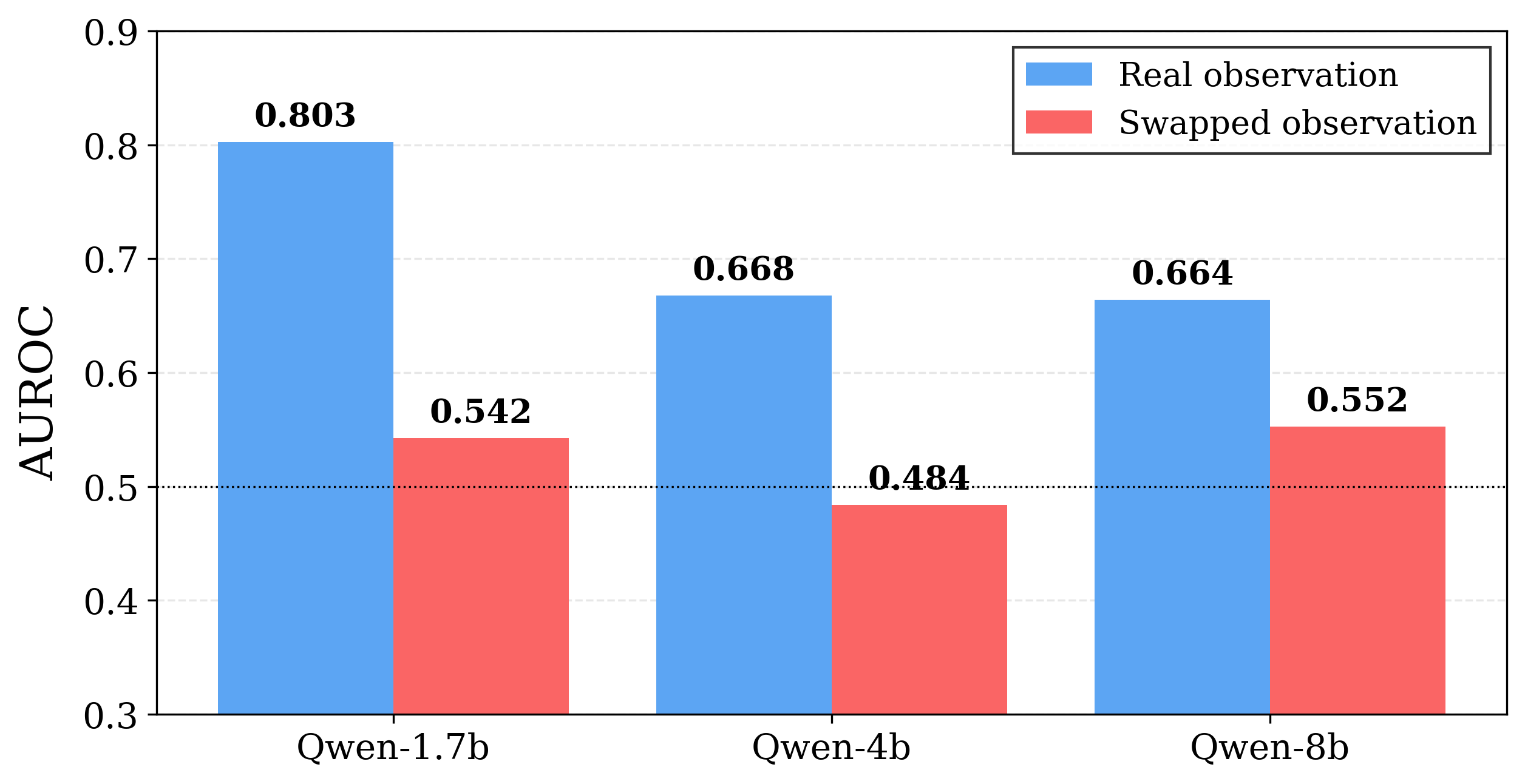}
    \caption{AUROC for distinguishing valid and invalid steps under realized and swapped future observations.}
    \label{fig:deep}
    \vspace{-2mm}
\end{figure}

\paragraph{Observation swapping analysis.} 
To examine whether the observation residual depends on the correspondence
between the future observation and the current action, we replace the
realized future observation in the Full view with a swapped observation
randomly sampled from another interaction step, while keeping all other
inputs unchanged. Following the step-level aggregation in Eq.~\ref{eq:step_level_score}, we set
the token-level signal to $\lvert e_{i,k,t} \rvert$ and use the resulting
step score to distinguish valid from invalid steps. As shown in
Fig.~\ref{fig:deep}, the AUROC values under the realized observations are 0.803,
0.668, and 0.664 across the three settings, respectively. After replacing
them with swapped observations, the corresponding AUROC values decrease
to 0.542, 0.484, and 0.552, all close to random ranking. These results
indicate that the discriminative ability of the residual primarily depends
on the correspondence between the future environment feedback and the
current action, rather than merely on the introduction of additional
observation text.

\paragraph{Computational Budget.}

\begin{figure}[t]
    \centering
    \includegraphics[width=\columnwidth]{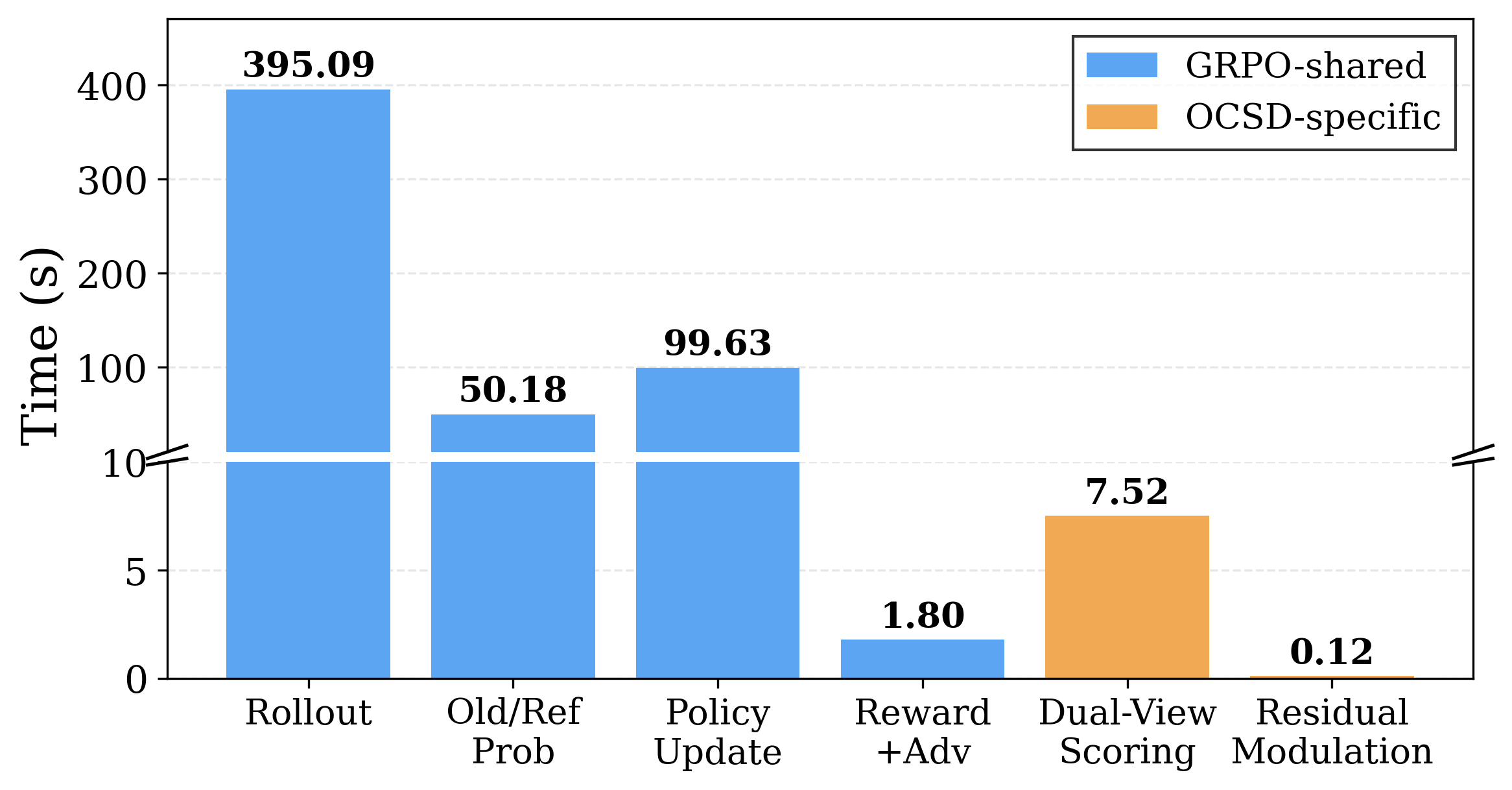}
    \caption{
    Per-iteration runtime breakdown of OCSD on ALFWorld with Qwen3-4B using 8 A100 GPUs. Blue bars denote modules shared with GRPO, while orange bars denote OCSD-specific operations. The y-axis uses a broken scale to accommodate small values.
    }
    \label{fig:runtime_breakdown}
    \vspace{-2mm}
\end{figure}

We record a detailed breakdown of the per-iteration training runtime
and decompose it into
modules shared with GRPO and OCSD-specific operations.
As shown in Figure~\ref{fig:runtime_breakdown}, Dual-view replay scoring and residual modulation require only 7.52 and 0.12
seconds per iteration, respectively. Together, they introduce an overhead of 7.64 seconds, corresponding to only 1.40\% relative to the shared GRPO pipeline. OCSD therefore maintains training efficiency comparable to standard GRPO.

\section{Related Work}
\paragraph{On-Policy Self-Distillation for Agentic Reinforcement Learning.}
OPSD provides fine-grained token-level supervision by re-evaluating
student-generated outputs under a privileged view~\citep{pidlm,opsd}.
Recent studies extend this paradigm to multi-turn agentic reinforcement
learning. RLSD uses token-level probability gaps to modulate update
magnitudes while retaining environment rewards to determine the update
direction~\citep{rlsd}. SDAR instead incorporates OPSD as a gated
auxiliary objective to improve training stability in multi-turn agent
settings~\citep{sdar}. Other work explores skill-conditioned agentic distillation through hierarchical skill extraction, joint skill-policy evolution, and local
selection between skill-conditioned and skill-free views
~\citep{opid,seed,ucob}, or derives privileged teacher context from
question-construction paths generated through self-play
~\citep{piplay}.

Recent studies also investigate step-level hindsight re-scoring
~\citep{stepopsd}, Bayes-calibrated turn-level credit derived from
verified answers~\citep{pbsd}, diagnostic feedback derived from
subsequent environment observations~\citep{hero}, and selective credit
adjustment using grounded environment feedback~\citep{serl}. These methods mainly study how privileged supervision is constructed and used, whereas we focus on disentangling future observations from shared replay scaffolds.

\paragraph{Calibration and Reliability of Privileged Signals}
Recent studies show that privileged distillation signals may contain
changes unrelated to the target information. RLCSD contrasts correct
and incorrect hints to reduce hint-induced style shifts
~\citep{rlcsd}. Purified OPSD uses a reference-only view to control
reference-specific shortcuts~\citep{purifiedopsd}, while AR-OPD
introduces a partially privileged anchor to constrain hindsight guidance
that is difficult for the current policy to reach~\citep{aropd}.
OPD$^2$ instead contrasts a reasoning-tuned teacher with its base model
to capture prediction changes associated with reasoning specialization
~\citep{opd2}. These approaches primarily address shifts induced by
hints, references, or teacher specialization in reasoning tasks.

Other studies analyze positional variation in teacher-token reliability
~\citep{pwopsd,iwopd}, the effects of privileged supervision on
high-entropy reasoning and self-correction
~\citep{rethinkingopsd,dasd}, the suppression of native reasoning under
overly strong supervision~\citep{adopsd,whydegrades}, and degenerate
response patterns induced by dense distillation, including formatting
artifacts and length-based shortcuts~\citep{denseropsd,demystifyingopd}. In contrast, we focus on separating future-observation effects from
shared replay scaffold effects in agentic replay signals.

\section{Conclusion}
We identify an attribution confound in agentic privileged replay scoring, where Full-view support contains score changes associated with both future observations and the shared replay scaffold. To address this issue, we propose OCSD, which derives an observation residual from structurally matched Full and Observation-Ablated replay views and uses it to calibrate token-level GRPO updates. Experiments across three agentic benchmarks and three Qwen3 model scales demonstrate consistent performance gains, while diagnostic analyses show that the calibrated residual better aligns with local environment feedback.

\bibliography{ref}


\clearpage
\appendix

\section{Additional Method Details}
\label{app:method}
\subsection{End-to-End Training Procedure}
\label{app:training_pipeline}

OCSD builds on the standard GRPO rollout-and-update pipeline and introduces
dual-view replay scoring between trajectory collection and policy
optimization. At each training iteration, the rollout policy first samples a
group of interaction trajectories for each task, and the terminal rewards are
used to compute the group-relative trajectory advantages. Following the
NLL-guided step-selection procedure described in
Section~\ref{sec:uncertainty_step_selection}, we compute the average token NLL
under the old policy for each interaction step, rank the steps within each
trajectory, and select the top-$\rho$ high-NLL steps.

For each selected step, OCSD constructs structurally matched Full and
Observation-Ablated replay views and re-scores all tokens generated at that
step under the same student-generated prefix. The difference between the
token scores assigned by the two views yields the observation residual, which
is then mapped to the bounded calibration signal $q_{i,k,t}$. This signal
modulates the magnitude of the original GRPO advantage at the token level
without changing the update direction determined by the trajectory-level
advantage. Tokens belonging to unselected steps retain the original
trajectory-level advantage.

OCSD introduces neither an additional imitation objective nor
replay-conditioned action generation. The Full and Observation-Ablated views
are used only during training to score tokens already generated by the
student, and the resulting observation residual is subsequently incorporated
into policy optimization.

\begin{algorithm}[t]
\caption{Observation-Calibrated Self-Distillation (OCSD)}
\label{alg:ocsd}
\footnotesize
\begin{algorithmic}[1]

\REQUIRE Policy $\pi_\theta$, reference policy $\pi_{\rm ref}$,
task distribution $\mathcal{D}$
\REQUIRE Group size $G$, selection ratio $\rho$,
modulation coefficient $\beta$

\FOR{each training iteration}
    \STATE Set $\theta_{\rm old}\leftarrow\theta$
    \STATE Sample a task batch $\mathcal{B}\sim\mathcal{D}$

    \STATE \textit{On-policy rollout}
    \FOR{each task $x\in\mathcal{B}$}
        \STATE Sample $G$ trajectories
        $\{\tau_i\}_{i=1}^{G}$ using
        $\pi_{\theta_{\rm old}}$
        \STATE Obtain terminal rewards
        $\{R(\tau_i)\}_{i=1}^{G}$
        \STATE Compute trajectory advantages
        $\{\widehat A_i\}_{i=1}^{G}$
    \ENDFOR

    \STATE \textit{NLL-guided step selection}
    \FOR{each sampled trajectory $\tau_i$}
        \STATE Compute average token NLL $u_{i,k}$
        for every interaction step
        \STATE Select the top-$\rho$ high-NLL steps
        $\mathcal{S}_i$

        \STATE \textit{Structurally matched replay scoring}
        \FOR{each selected step $k\in\mathcal{S}_i$}
            \STATE Construct Full evidence $E^F_{i,k}$ and
            Observation-Ablated evidence $E^A_{i,k}$
            \STATE Score all generated tokens under the Student,
            Full, and Observation-Ablated views using
            $\pi_{\theta_{\rm old}}$
            \STATE Compute $\delta^F_{i,k,t}$,
            $\delta^A_{i,k,t}$, $e_{i,k,t}$, and $q_{i,k,t}$
        \ENDFOR

        \STATE Construct
        $\widehat A^{\rm OCSD}_{i,k,t}$
        using Eq.~\eqref{eq:ocsd_advantage}
    \ENDFOR

    \STATE Update $\theta$ by maximizing
    $\mathcal{J}^{\rm OCSD}(\theta)$ in
    Eq.~\eqref{eq:ocsd_objective}
\ENDFOR

\ENSURE Optimized policy $\pi_\theta$

\end{algorithmic}
\end{algorithm}

The complete OCSD training procedure is summarized in
Algorithm~\ref{alg:ocsd}, with the construction of replay evidence
and replay-scoring details described in
Sections~\ref{app:replay_evidence} and~\ref{app:replay_scoring},
respectively.

\subsection{Construction of Replay Evidence}
\label{app:replay_evidence}

For each selected interaction step, OCSD constructs two structurally matched
replay views, Full and Observation-Ablated. In all main experiments, the
replay evidence uses a two-observation horizon consisting of the two
post-action observations $o_{k+1}$ and $o_{k+2}$, together with a naturalized
schema of the intervening action $a_{k+1}$. Accordingly, the Full evidence
$E^F_{i,k}$ contains
$o_{k+1},\quad g(a_{k+1}),\quad o_{k+2}$,
where $g(\cdot)$ converts a future action into a natural-language schema that
preserves its coarse intent and syntactic structure while replacing
high-information arguments, such as concrete entities, locations, queries,
and answers, with their coarse semantic roles. For example,
\texttt{take soapbar 1 from countertop 1} is converted to
\texttt{take an item from a receptacle}, while
\texttt{Search[Barack Obama birth year]} is converted to
\texttt{Search[a query]}. When a future action cannot be reliably parsed, we
use the fixed fallback schema \texttt{a future action}.

The Observation-Ablated evidence $E^A_{i,k}$ preserves the same field order,
replay format, and future-action scaffold as the Full evidence, but replaces
each future-observation field with the fixed phrase
\texttt{Observation: not provided}. The two replay views therefore differ
only in whether the actual future observations are present, while the
remaining replay context is kept unchanged.

At a trajectory boundary where no subsequent action is available, the replay
evidence contains only the available future observation following the current
action. The Full view retains its actual content, whereas the
Observation-Ablated view uses the fixed replacement phrase at the same
position without introducing an additional future-action scaffold.

\subsection{Replay Scoring and Optimization Details}
\label{app:replay_scoring}

For each selected interaction step, OCSD uses the Student, Full, and
Observation-Ablated views to score the same tokens already generated by the
student at that step. The three views use the same model parameters
$\theta_{\mathrm{old}}$ but differ in their input contexts. The Student view
conditions only on the original interaction context $h_{i,k,t}$. The Full
view additionally conditions on the Full evidence $E^F_{i,k}$, whereas the
Observation-Ablated view conditions on the structurally matched
Observation-Ablated evidence $E^A_{i,k}$. In all three views, the context
preceding the scored token contains the same student-generated prefix
$y_{i,k,<t}$. The Full and Observation-Ablated views therefore perform
replay-conditioned scoring of tokens already generated during rollout,
without generating an alternative response, action, or trajectory.

Following the definition of privileged replay scoring in the main text, the
Full support $\delta^F_{i,k,t}$ and Observation-Ablated support
$\delta^A_{i,k,t}$ measure the token log-probability changes induced by their
respective replay evidence relative to the Student view. OCSD computes their
difference as
\begin{equation}
\label{eq:residual_support_appendix}
e_{i,k,t}
=
\delta^F_{i,k,t}
-
\delta^A_{i,k,t}.
\end{equation}
Because the two support terms share the same Student-view term, this common
term cancels in their difference. The observation residual can therefore
also be written as
\begin{equation}
\label{eq:residual_scores_appendix}
\begin{aligned}
e_{i,k,t}
={}&
\log \pi_F
\left(
y_{i,k,t}
\mid
h_{i,k,t}, E^F_{i,k}
\right)
\\
&-
\log \pi_A
\left(
y_{i,k,t}
\mid
h_{i,k,t}, E^A_{i,k}
\right).
\end{aligned}
\end{equation}
This difference contrasts how the two structurally matched replay views
score the same generated token. It discounts score changes shared by the
replay scaffold and yields a token-level signal that is more sensitive to
the actual future observation. OCSD then maps the observation residual to
the bounded calibration signal $q_{i,k,t}$ following the definition in the
main text.

During policy optimization, $q_{i,k,t}$ is used neither as an independent
reward nor as an imitation target. Instead, it modulates the magnitude of the
original trajectory-level advantage at the token level. The sign of the
trajectory-level advantage continues to determine the update direction,
whereas the observation residual only adjusts the update strength assigned
to each token. This calibration is applied to all generated tokens within
each selected step rather than only to action tokens. Steps not selected by
the NLL-guided step-selection procedure are not subjected to Full and
Observation-Ablated replay scoring, and all of their tokens retain the
original trajectory-level advantage.

\section{Experimental Setup Details}
\label{app:experimental_setup}

\begin{table}[tbp]
\centering
\small
\setlength{\tabcolsep}{4pt}
\begin{tabular}{@{}p{0.48\linewidth}p{0.47\linewidth}@{}}
\toprule
\textbf{Hyperparameter} & \textbf{Value} \\
\midrule

\multicolumn{2}{@{}l}{\textbf{RL Training}} \\

Backbone models
& Qwen3-1.7B / Qwen3-4B / Qwen3-8B \\

Optimizer
& AdamW \\

Learning rate
& $1\times10^{-6}$ \\

Weight decay
& 0.01 \\

Rollout group size $G$
& 8 \\

GRPO clipping coefficient $\epsilon$
& 0.2 \\

Maximum response length
& 512 \\

Rollout temperature
& 1.0 \\

Rollout top-$p$
& 1.0 \\

Training steps
& 150 \\

Gradient clipping
& 1.0 \\

Training hardware
& 8 NVIDIA A100 GPUs \\

\midrule
\multicolumn{2}{@{}l}{\textbf{OCSD}} \\

Step selection ratio $\rho$
& 0.2 \\

Advantage modulation coefficient $\beta$
& 0.5 \\

\bottomrule
\end{tabular}
\caption{
Shared hyperparameters used for RL training and OCSD across all benchmarks.
}
\label{tab:shared_training_hyperparameters}
\end{table}

\begin{table}[tbp]
\centering
\small
\setlength{\tabcolsep}{3pt}
\begin{tabular}{@{}p{0.34\linewidth}ccc@{}}
\toprule
\textbf{Hyperparameter}
& \textbf{ALFWorld}
& \textbf{WebShop}
& \textbf{\shortstack{Search-\\QA}} \\
\midrule

Task batch size
& 16
& 16
& 128 \\

KL coefficient $\beta_{\mathrm{KL}}$
& 0.01
& 0.01
& 0.001 \\

Maximum prompt length
& 2048
& 4096
& 4096 \\

Maximum interaction steps
& 50
& 15
& 4 \\

\bottomrule
\end{tabular}
\caption{
Benchmark-specific training configurations. Task batch size denotes the
number of tasks sampled in each training batch, with $G=8$ rollout
trajectories generated for each task.
}
\label{tab:benchmark_training_hyperparameters}
\end{table}

\paragraph{Training Protocol.}
We conduct experiments with Qwen3-1.7B, Qwen3-4B, and Qwen3-8B.
Following SDAR~\cite{sdar}, we adopt a unified training protocol for all
post-training methods. Shared optimization and rollout hyperparameters are
kept consistent across all post-training methods. For published baselines,
we follow the method-specific configurations reported in the corresponding
original papers.

All three benchmarks require the model to produce an explicit reasoning
segment before each action. However, Qwen3's native thinking mode already
generates a reasoning block delimited by \texttt{<think>} and
\texttt{</think>}. Enabling this mode together with the benchmark-specific
output format can therefore produce duplicated or nested thinking delimiters,
leading to ambiguous output parsing. We consequently disable the native
thinking mode by setting \texttt{enable\_thinking=False} and instruct the
model to place its intermediate reasoning within
\texttt{<reason>} and \texttt{</reason>} tags before emitting the action.

For ALFWorld and WebShop, each training batch contains
16 tasks, with $G=8$ rollout trajectories sampled for each task. For
Search-QA, each training batch contains 128 tasks with the same rollout group
size. All post-training methods are trained for 150 optimization steps.

Except for the benchmark-specific configurations, all benchmarks and model
scales share the same optimizer, learning rate, weight decay, GRPO clipping
coefficient, rollout decoding parameters, maximum response length, and
gradient clipping. All experiments are conducted using 8 NVIDIA A100 80GB GPUs.
The shared RL training and OCSD hyperparameters are summarized in
Table~\ref{tab:shared_training_hyperparameters}, while the benchmark-specific
task batch sizes, KL coefficients, maximum prompt lengths, and maximum
interaction steps are reported in
Table~\ref{tab:benchmark_training_hyperparameters}.

For OCSD, we use a step selection ratio of $\rho=0.2$ and an advantage
modulation coefficient of $\beta=0.5$ across all benchmarks and model scales.
These two hyperparameters are not adjusted separately for individual
benchmarks or model sizes.

\paragraph{Evaluation Protocol.}
We use the same evaluation tasks and inference configuration for all methods.
For both ALFWorld and WebShop, one interaction trajectory is generated for
each evaluation task, with a maximum of 50 interaction steps for ALFWorld and
15 interaction steps for WebShop. For ALFWorld, we report the success rates
for the six household manipulation task categories together with the overall
result. For WebShop, we report both the task score returned by the environment
and the success rate.

For Search-QA, we follow the same evaluation protocol as
SDAR~\cite{sdar} and evaluate each of the seven datasets using its
corresponding test or validation split. NQ and HotpotQA are treated as the
in-domain datasets, while the remaining five datasets are used for
out-of-domain evaluation. For each question, the agent generates one
trajectory with at most four interaction steps. Evaluation uses a validation
batch size of 512. We report Exact Match (EM), micro-averaged over the
in-domain datasets (ID), the out-of-domain datasets (OOD), and all seven
datasets (Avg.).

\section{Details of Fine-Grained Diagnosis}
\label{app:fine_grained_diagnosis}

\subsection{Word-Level Preference Analysis}
\label{app:word_preference}

Figure~\ref{fig:token_diagnosis}(a) analyzes only the top-$20\%$
high-NLL interaction steps selected by OCSD within each trajectory.
Unselected steps are excluded. Since the model tokenizer may split one word
into multiple byte-pair encoding (BPE) subword tokens, we merge adjacent
subword pieces into word occurrences using whitespace and structural
boundaries. The signal of each word occurrence is the mean of its constituent
token signals. Repeated occurrences of the same word are then averaged within
each checkpoint, followed by equal-weight averaging across checkpoints.

Within each checkpoint, we separately center $\delta^F$, $\delta^A$, and
$e=\delta^F-\delta^A$. Full+, Ablated+, and Residual+ denote occurrences in
the upper $5\%$ tail of the corresponding centered signal, while Residual$-$
denotes the lower $5\%$ tail. Thus, the ``+'' and ``$-$'' labels indicate
checkpoint-relative tail membership rather than the sign of the raw signal.

For each word $w$, the tail-hit rate is the fraction of its occurrences that
enter the corresponding tail. Its displayed word-cloud size is determined by
\[
\operatorname{tail\_ratio}(w)
\log\!\left(1+\operatorname{count}(w)\right),
\]
which balances tail preference with log-scaled occurrence frequency.

\subsection{Functional-Role Categorization}
\label{app:functional_categories}

Figure~\ref{fig:token_diagnosis}(b) uses the same top-$20\%$ high-NLL
steps as Figure~\ref{fig:token_diagnosis}(a). Each word is automatically
assigned to one of five mutually exclusive categories using deterministic
lexical and output-structure rules, with priority
\[
\text{Format}>\text{Action}>\text{Entity/State}>
\text{Reasoning}>\text{Other}.
\]
Format includes structural tags and output-template markers. Action includes
verbs such as \texttt{go}, \texttt{take}, \texttt{put}, and
\texttt{examine}. Entity/State includes environment entities, receptacles,
and state words. Reasoning includes connectives such as \texttt{because},
\texttt{since}, and \texttt{however}, while all remaining words are assigned
to Other.

To avoid ambiguity with the GRPO clipping coefficient $\epsilon$, we denote
the numerical stabilizer in Eq.~\eqref{eq:relative_residual_contribution} by
$\epsilon_D$ in this appendix. For each word occurrence, we compute
\[
D_{i,k,w}
=
\frac{m^R_{i,k,w}}
{m^R_{i,k,w}+m^A_{i,k,w}+\epsilon_D},
\]
where $m^R_{i,k,w}$ and $m^A_{i,k,w}$ are the mean absolute observation
residual and Observation-Ablated support over its constituent tokens,
respectively. Here, $\epsilon_D>0$ is a small numerical stabilizer distinct
from the GRPO clipping coefficient. Accordingly, $D\in[0,1)$ measures the relative
contribution of the residual magnitude. We estimate the mean and 95\% confidence interval using
1,000 trajectory-cluster bootstrap samples and aggregate checkpoints with
equal weight.

\subsection{Local-Feedback Labels and AUROC}
\label{app:local_feedback}

Figure~\ref{fig:token_diagnosis}(c) analyzes only the top-$20\%$ high-NLL
steps selected by OCSD. We automatically assign labels using the environment
execution status and the immediate future observation. A step is labeled
invalid if the action is rejected by the environment or if the feedback
contains an explicit error or no-effect template, including
\texttt{nothing happens}, \texttt{cannot}, \texttt{not carrying},
\texttt{no such}, \texttt{invalid}, or \texttt{nothing to}. A step is labeled
valid if the action is accepted, produces environment progress, and does not
contain a no-effect template. All remaining ambiguous cases are excluded.

For each interaction step, we average the signal over all generated tokens.
The residual score uses the signed value
$e=\delta^F-\delta^A$, rather than its magnitude. We pool the valid and
invalid steps across checkpoints and evaluate each signal using AUROC.
We report 95\% confidence intervals from 1,000 trajectory-cluster bootstrap
samples, where complete trajectories, rather than individual steps, are
resampled.

\subsection{Robustness Settings}
\label{app:diagnostic_robustness}

Figure~\ref{fig:token_diagnosis}(d) reports the AUROC improvement of the
residual over Full support:
\[
\Delta\mathrm{AUROC}
=
\mathrm{AUROC}(\mathrm{Residual})
-
\mathrm{AUROC}(\mathrm{Full}).
\]

Raw directly uses the original step-level Full and residual scores, obtained
by averaging token signals within each interaction step, and computes AUROC
over the pooled checkpoints. Standard separately applies within-checkpoint
z-score normalization to the Full and residual scores before pooling, thereby
reducing scale drift across training.

Matched applies stratified class balancing together with checkpoint-balanced
aggregation. We stratify samples by the joint combination of training stage
and action operator, where training is divided into early (steps 1--50),
middle (steps 51--100), and late (steps 101--150) stages. The action operator
is extracted from the verb prefix of the executed action, with actions outside
the predefined operators assigned to \texttt{other}. Within each stratum,
valid and invalid samples are randomly downsampled to the same size, with at
most 50 samples retained from each class. AUROC is then computed separately
for each checkpoint and averaged equally across checkpoints, preventing
checkpoints with more samples from dominating the result.

All settings use 1,000 paired trajectory-cluster bootstrap samples. Full and
residual scores are evaluated on the same resampled trajectories, and 95\%
confidence intervals are computed directly from their paired AUROC
differences.

\section{Observation-Swapping Sensitivity Analysis}
\label{app:observation_swapping}

To examine whether the observation residual depends on the correspondence
between the current action and its immediate post-action observation
$o_{k+1}$, we conduct an observation-swapping analysis using OCSD actor
checkpoints of Qwen3-1.7B, Qwen3-4B, and Qwen3-8B. For each model, we select checkpoints at training
steps 10, 50, 100, and 150 to cover different stages of training. The
evaluation samples are drawn from the high-NLL steps selected by OCSD at each
checkpoint and are stratified into three feedback categories: progress,
no-progress, and invalid. We sample at most 40 steps from each category at
each checkpoint without reusing trajectories. The resulting evaluation set
contains 560 steps, including 208, 186, and 166
steps for Qwen3-1.7B, Qwen3-4B, and Qwen3-8B, respectively.

For each target step, we independently sample five donor observations from
steps in different trajectories whose actions share the same operator as the
target action. Each donor observation is the immediate post-action
observation returned after the corresponding donor action. For the $j$-th
donor, we replace only the realized observation $o_{k+1}$ in the Full view
with the donor observation, while keeping the original future-action schema
$g(a_{k+1})$, the second future observation $o_{k+2}$, the student output,
the replay format, and the Observation-Ablated view unchanged. Both the
realized and swapped views are scored by the corresponding actor checkpoint
using the same response token sequence. To reduce variance associated with
any single donor observation, the final swapped score averages the
step-level scores obtained from the five donor observations.

For the realized immediate observation and the $j$-th donor observation, we
respectively compute
\[
\begin{aligned}
e^{\mathrm{real}}_{i,k,t}
={}&
\log \pi_F
\left(
y_{i,k,t}
\mid
h_{i,k,t},
E^{F,\mathrm{real}}_{i,k}
\right)
\\
&-
\log \pi_A
\left(
y_{i,k,t}
\mid
h_{i,k,t},
E^A_{i,k}
\right),
\end{aligned}
\]
and
\[
\begin{aligned}
e^{\mathrm{swap},j}_{i,k,t}
={}&
\log \pi_F
\left(
y_{i,k,t}
\mid
h_{i,k,t},
E^{F,\mathrm{swap},j}_{i,k}
\right)
\\
&-
\log \pi_A
\left(
y_{i,k,t}
\mid
h_{i,k,t},
E^A_{i,k}
\right).
\end{aligned}
\]
Here, $E^{F,\mathrm{swap},j}_{i,k}$ differs from
$E^{F,\mathrm{real}}_{i,k}$ only in the immediate observation field:
$o_{k+1}$ is replaced by the $j$-th donor observation, while
$g(a_{k+1})$ and $o_{k+2}$ remain unchanged.

The two residuals share the same Observation-Ablated support, and their Full
views differ only in the immediate observation $o_{k+1}$. Their difference
therefore arises only from the immediate post-action observation provided to
the Full teacher.

We measure observation sensitivity using the mean residual magnitude over all
generated tokens in a step. For the realized observation, the score is
defined as
\[
S^{\mathrm{real}}_{i,k}
=
\frac{1}{T_{i,k}}
\sum_{t=1}^{T_{i,k}}
\left|e^{\mathrm{real}}_{i,k,t}\right|,
\]
while the swapped score is
\[
S^{\mathrm{swap}}_{i,k}
=
\frac{1}{5}
\sum_{j=1}^{5}
\frac{1}{T_{i,k}}
\sum_{t=1}^{T_{i,k}}
\left|e^{\mathrm{swap},j}_{i,k,t}\right|.
\]
Thus, we first compute a step-level residual magnitude for each donor
observation and then average the scores across the five donor observations.
We use these scores to distinguish valid steps that produce environment
progress from invalid steps and compute AUROC under the realized and swapped
observations. No-progress steps are excluded from this binary evaluation.

This metric addresses a different question from the signed residual used in
Figure~\ref{fig:token_diagnosis}(c). Figure~\ref{fig:token_diagnosis}(c)
examines whether the direction of the residual aligns with local environment
feedback. In contrast, observation swapping evaluates the response strength
of the Full teacher relative to the Observation-Ablated teacher and whether
the feedback-discriminative information in this response depends on the
correct observation--action correspondence. We therefore use $\lvert e\rvert$
to construct the observation-sensitivity score and compare AUROC under
realized and swapped observations. The two analyses respectively measure
directional feedback alignment and observation-dependent sensitivity.

As shown in Figure~\ref{fig:deep}, the realized immediate post-action
observations consistently yield higher AUROC across all three model scales.
Replacing only $o_{k+1}$ with donor observations from different trajectories
whose actions share the same operator reduces AUROC from 0.803, 0.668, and
0.664 to 0.542, 0.484, and 0.552 for Qwen3-1.7B, Qwen3-4B, and Qwen3-8B,
respectively. These results indicate that the local feedback information
captured by the residual depends on the correct correspondence between the
immediate post-action observation and the current interaction step, rather
than merely on the presence of arbitrary additional observation text in the
Full view.

\section{Hyperparameter Sensitivity}

\begin{figure*}[t]
    \centering
    \includegraphics[width=\textwidth]{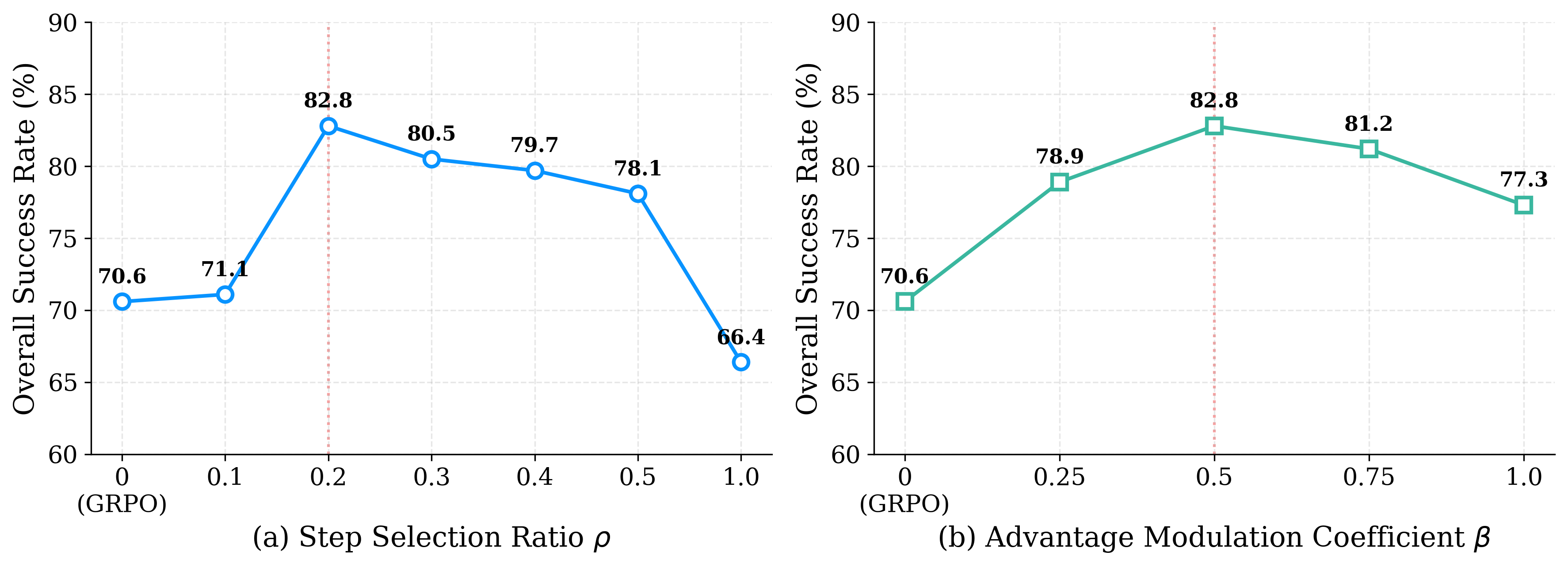}
    \caption{
    Hyperparameter sensitivity of OCSD with Qwen3-4B on ALFWorld.
    (a) Effect of the step selection ratio $\rho$ with the advantage
    modulation coefficient fixed at $\beta=0.5$.
    (b) Effect of the advantage modulation coefficient $\beta$ with the
    step selection ratio fixed at $\rho=0.2$.
    For visualization, $\rho=0$ and $\beta=0$ denote the GRPO baseline
    without residual calibration. The red dotted lines indicate the
    default settings used in the main experiments.
    }
    \label{fig:hyperparameter_sensitivity}
\end{figure*}

We examine the sensitivity of OCSD to the step selection ratio $\rho$ and
the advantage modulation coefficient $\beta$ using Qwen3-4B on ALFWorld.
For each analysis, we vary one hyperparameter while fixing the other to its
default value. As shown in Figure~\ref{fig:hyperparameter_sensitivity}(a),
selecting only $10\%$ of the interaction steps yields limited improvement
over GRPO, whereas increasing the ratio to $\rho=0.2$ raises the overall
success rate from $70.6$ to $82.8$. Performance then gradually decreases as
more steps are selected, reaching $66.4$ when residual calibration is
applied to all interaction steps at $\rho=1.0$. This trend suggests that
concentrating the observation-calibrated token signal on a moderate fraction
of high-NLL steps is important. Selecting too few steps provides insufficient
coverage, while calibrating all steps may introduce low-value or noisy
signals. We therefore use $\rho=0.2$ in the main experiments.

Figure~\ref{fig:hyperparameter_sensitivity}(b) shows a similar pattern for
the advantage modulation coefficient. Setting $\beta=0$ removes residual
calibration and recovers the GRPO baseline. Increasing $\beta$ to $0.25$
and $0.5$ improves the overall success rate to $78.9$ and $82.8$,
respectively. Performance remains strong at $\beta=0.75$ but decreases to
$77.3$ at $\beta=1.0$. These results indicate that moderate modulation
strength makes effective use of the observation-calibrated token signal,
whereas overly strong modulation may place excessive weight on the residual.
Accordingly, we set $\beta=0.5$ throughout the main experiments.

\section{Additional Experimental Results}
\subsection{Full Search-QA Results}
\begin{table*}[!t]
\centering

{
\small
\setlength{\tabcolsep}{1pt}

\begin{tabular*}{\textwidth}{
@{\extracolsep{\fill}}
l
cc
@{\hspace{4pt}}
ccccc
@{\hspace{4pt}}
c
@{}
}
\toprule
\multirow{2}{*}{\textbf{Method}} &
\multicolumn{2}{c}{\textbf{ID}} &
\multicolumn{5}{c}{\textbf{OOD}} &
\multirow{2}{*}{\textbf{Avg.}} \\
\cmidrule(lr){2-3}
\cmidrule(lr){4-8}
&
\textbf{NQ} &
\textbf{HotpotQA} &
\textbf{TriviaQA} &
\textbf{PopQA} &
\textbf{2Wiki} &
\textbf{MuSiQue} &
\textbf{Bamboogle} &
\\
\midrule

\rowcolor[gray]{.95}
\multicolumn{9}{l}{\textit{Qwen3-1.7B}} \\

Vanilla
& 22.9
& 18.0
& 33.4
& 22.6
& 19.0
& 3.7
& 5.6
& 22.5 \\

GRPO
& \meanstd{40.1}{0.3}
& \meanstd{34.4}{2.3}
& {\cellcolor{second}\meanstd{58.4}{0.5}}
& \meanstd{44.8}{1.5}
& \meanstd{30.3}{0.4}
& \meanstd{10.8}{2.4}
& \meanstd{64.6}{1.2}
& \meanstd{40.9}{0.7} \\

OPSD
& \meanstd{1.0}{1.3}
& \meanstd{1.6}{2.0}
& \meanstd{2.6}{3.1}
& \meanstd{1.4}{1.6}
& \meanstd{4.1}{4.8}
& \meanstd{0.2}{0.2}
& \meanstd{0.2}{0.2}
& \meanstd{2.3}{2.7} \\

GRPO+OPSD
& {\cellcolor{second}\meanstd{41.1}{1.9}}
& {\cellcolor{second}\meanstd{35.8}{2.3}}
& {\cellcolor{best}\bfseries\boldmath\meanstd{58.8}{0.6}}
& {\cellcolor{second}\meanstd{46.6}{0.9}}
& \meanstd{33.7}{1.9}
& {\cellcolor{second}\meanstd{12.3}{1.2}}
& {\cellcolor{second}\meanstd{66.0}{0.8}}
& {\cellcolor{second}\meanstd{42.6}{0.4}} \\

RLSD
& \meanstd{39.1}{1.9}
& \meanstd{35.5}{0.4}
& \meanstd{57.2}{0.4}
& \meanstd{44.1}{0.8}
& \meanstd{32.2}{1.6}
& {\cellcolor{second}\meanstd{12.3}{1.3}}
& {\cellcolor{best}\bfseries\boldmath\meanstd{66.5}{0.8}}
& \meanstd{41.0}{0.7} \\

SDAR
& \meanstd{38.8}{0.8}
& \meanstd{35.7}{0.4}
& \meanstd{57.8}{1.1}
& \meanstd{44.7}{0.6}
& {\cellcolor{second}\meanstd{34.5}{1.2}}
& {\cellcolor{best}\bfseries\boldmath\meanstd{12.5}{0.3}}
& \meanstd{65.3}{0.4}
& \meanstd{41.8}{0.8} \\

\textbf{OCSD (ours)}
& {\cellcolor{best}\bfseries\boldmath\meanstd{41.2}{2.1}}
& {\cellcolor{best}\bfseries\boldmath\meanstd{35.9}{0.5}}
& {\cellcolor{second}\meanstd{58.4}{0.7}}
& {\cellcolor{best}\bfseries\boldmath\meanstd{47.1}{1.0}}
& {\cellcolor{best}\bfseries\boldmath\meanstd{35.1}{2.2}}
& \meanstd{11.9}{1.0}
& \meanstd{65.7}{0.5}
& {\cellcolor{best}\bfseries\boldmath\meanstd{43.1}{1.2}} \\

\midrule

\rowcolor[gray]{.95}
\multicolumn{9}{l}{\textit{Qwen3-4B}} \\

Vanilla
& 30.5
& 25.8
& 49.5
& 32.5
& 28.7
& 7.3
& 14.1
& 33.0 \\

GRPO
& {\cellcolor{second}\meanstd{45.9}{0.8}}
& \meanstd{40.2}{0.5}
& \meanstd{62.5}{0.4}
& {\cellcolor{second}\meanstd{46.8}{1.4}}
& \meanstd{37.0}{2.4}
& \meanstd{16.0}{1.3}
& \meanstd{71.0}{0.8}
& \meanstd{45.5}{0.8} \\

OPSD
& \meanstd{0.7}{0.9}
& \meanstd{0.1}{0.1}
& \meanstd{0.9}{1.3}
& \meanstd{1.3}{1.7}
& \meanstd{0.1}{0.1}
& \meanstd{0.0}{0.0}
& \meanstd{0.0}{0.0}
& \meanstd{0.6}{0.8} \\

GRPO+OPSD
& \meanstd{45.3}{0.8}
& {\cellcolor{second}\meanstd{41.1}{1.4}}
& \meanstd{62.2}{0.7}
& \meanstd{45.4}{1.1}
& \meanstd{38.4}{2.6}
& \meanstd{16.9}{1.2}
& \meanstd{70.6}{1.1}
& \meanstd{45.5}{1.2} \\

RLSD
& {\cellcolor{best}\bfseries\boldmath\meanstd{46.4}{1.0}}
& {\cellcolor{best}\bfseries\boldmath\meanstd{41.2}{0.2}}
& \meanstd{62.2}{0.5}
& {\cellcolor{second}\meanstd{46.8}{1.7}}
& \meanstd{39.8}{1.0}
& {\cellcolor{second}\meanstd{17.1}{1.8}}
& {\cellcolor{best}\bfseries\boldmath\meanstd{72.7}{0.2}}
& {\cellcolor{second}\meanstd{46.3}{0.5}} \\

SDAR
& \meanstd{45.7}{0.8}
& \meanstd{40.6}{0.2}
& {\cellcolor{second}\meanstd{62.8}{0.8}}
& \meanstd{45.2}{0.1}
& {\cellcolor{second}\meanstd{40.1}{0.2}}
& {\cellcolor{best}\bfseries\boldmath\meanstd{17.4}{1.1}}
& {\cellcolor{second}\meanstd{71.4}{1.1}}
& \meanstd{45.9}{0.2} \\

\textbf{OCSD (ours)}
& \meanstd{45.4}{0.1}
& {\cellcolor{second}\meanstd{41.1}{1.3}}
& {\cellcolor{best}\bfseries\boldmath\meanstd{63.2}{1.1}}
& {\cellcolor{best}\bfseries\boldmath\meanstd{48.2}{0.5}}
& {\cellcolor{best}\bfseries\boldmath\meanstd{42.8}{2.0}}
& \meanstd{17.0}{1.0}
& \meanstd{70.6}{1.8}
& {\cellcolor{best}\bfseries\boldmath\meanstd{47.5}{0.8}} \\

\midrule

\rowcolor[gray]{.95}
\multicolumn{9}{l}{\textit{Qwen3-8B}} \\

Vanilla
& 28.9
& 29.2
& 55.4
& 32.2
& 32.4
& 8.7
& 66.9
& 35.6 \\

GRPO
& \meanstd{46.2}{0.2}
& \meanstd{42.5}{0.1}
& \meanstd{64.2}{0.1}
& \meanstd{48.6}{0.5}
& {\cellcolor{second}\meanstd{45.1}{1.2}}
& \meanstd{17.8}{0.8}
& \meanstd{69.9}{1.0}
& {\cellcolor{second}\meanstd{48.7}{0.3}} \\

OPSD
& \meanstd{35.7}{0.4}
& \meanstd{32.7}{1.0}
& \meanstd{56.4}{1.3}
& \meanstd{41.2}{2.7}
& \meanstd{32.3}{1.7}
& \meanstd{12.5}{3.9}
& \meanstd{68.8}{3.5}
& \meanstd{39.5}{0.6} \\

GRPO+OPSD
& \meanstd{46.3}{0.8}
& {\cellcolor{second}\meanstd{43.2}{2.2}}
& {\cellcolor{second}\meanstd{64.8}{0.8}}
& \meanstd{47.8}{0.2}
& \meanstd{42.8}{1.8}
& \meanstd{17.9}{1.7}
& {\cellcolor{second}\meanstd{72.6}{1.6}}
& \meanstd{48.2}{0.2} \\

RLSD
& {\cellcolor{best}\bfseries\boldmath\meanstd{46.9}{0.7}}
& {\cellcolor{best}\bfseries\boldmath\meanstd{45.0}{0.6}}
& \meanstd{64.3}{0.8}
& {\cellcolor{second}\meanstd{48.8}{0.8}}
& \meanstd{41.0}{0.3}
& {\cellcolor{best}\bfseries\boldmath\meanstd{21.2}{1.3}}
& {\cellcolor{best}\bfseries\boldmath\meanstd{72.9}{0.6}}
& \meanstd{48.4}{0.1} \\

SDAR
& \meanstd{45.4}{0.6}
& \meanstd{43.0}{1.0}
& \meanstd{63.4}{0.6}
& \meanstd{47.1}{1.1}
& \meanstd{40.2}{2.3}
& {\cellcolor{second}\meanstd{19.0}{2.1}}
& \meanstd{71.5}{0.6}
& \meanstd{47.0}{0.4} \\

\textbf{OCSD (ours)}
& {\cellcolor{second}\meanstd{46.4}{0.8}}
& \meanstd{42.3}{0.5}
& {\cellcolor{best}\bfseries\boldmath\meanstd{65.2}{0.6}}
& {\cellcolor{best}\bfseries\boldmath\meanstd{49.0}{0.4}}
& {\cellcolor{best}\bfseries\boldmath\meanstd{45.6}{0.9}}
& \meanstd{17.0}{0.4}
& \meanstd{71.4}{1.1}
& {\cellcolor{best}\bfseries\boldmath\meanstd{49.1}{0.3}} \\

\bottomrule
\end{tabular*}
}

\caption{
Detailed performance on the two in-domain (ID) and five out-of-domain (OOD)
Search-QA datasets, together with the overall micro-averaged score across all
seven datasets. Results are reported as mean $\pm$ standard deviation over
three runs with different random seeds for most experiments. Blue and red
backgrounds denote the best and second-best methods, respectively.
}
\label{tab:search_qa_details}
\end{table*}

Table~\ref{tab:search_qa_details} presents the complete Search-QA results
across the seven constituent datasets for all three Qwen3 model scales,
including the two in-domain datasets used for training and the five
out-of-domain datasets used only for evaluation, together with the overall
micro-averaged score across all seven datasets.

\subsection{Detailed Ablation Results}

\begin{table*}[t]
\centering
\small
\setlength{\tabcolsep}{2.8pt}

\begin{tabular*}{\textwidth}{
@{\extracolsep{\fill}}
l
ccccccc
@{\hspace{4pt}}
cc
@{}
}
\toprule
\multirow{2}{*}{\textbf{Method}} &
\multicolumn{7}{c}{\textbf{ALFWorld}} &
\multicolumn{2}{c}{\textbf{WebShop}} \\
\cmidrule(lr){2-8}
\cmidrule(lr){9-10}
&
\textbf{Pick} &
\textbf{Look} &
\textbf{Clean} &
\textbf{Heat} &
\textbf{Cool} &
\textbf{Pick2} &
\textbf{Overall} &
\textbf{Score} &
\textbf{Success} \\
\midrule

GRPO
& \meanstd{86.6}{0.6}
& \meanstd{59.5}{5.1}
& \meanstd{77.7}{4.7}
& \meanstd{67.1}{4.1}
& \meanstd{55.5}{4.1}
& \meanstd{55.0}{6.2}
& \meanstd{70.6}{1.3}
& \meanstd{81.9}{3.6}
& \meanstd{69.5}{1.5} \\

\textit{w/o Ablated Teacher}
& \meanstd{88.6}{2.4}
& \meanstd{47.6}{3.4}
& \meanstd{81.1}{2.1}
& \meanstd{69.7}{3.9}
& \meanstd{77.2}{3.7}
& \meanstd{62.9}{1.0}
& \meanstd{74.8}{1.6}
& \meanstd{85.0}{1.0}
& \meanstd{72.9}{1.8} \\

\textit{w/ Random Selection}
& \meanstd{85.6}{2.4}
& \meanstd{50.0}{4.6}
& \meanstd{83.6}{4.2}
& \meanstd{83.1}{2.2}
& \bfseries\boldmath\meanstd{82.8}{4.2}
& \bfseries\boldmath\meanstd{68.2}{4.1}
& \meanstd{78.9}{1.1}
& \meanstd{84.2}{1.6}
& \meanstd{70.6}{2.6} \\

\textit{w/o Step Selection}
& \meanstd{82.2}{3.2}
& \meanstd{75.9}{1.3}
& \meanstd{58.7}{1.9}
& \meanstd{68.7}{2.8}
& \meanstd{41.8}{3.8}
& \meanstd{56.7}{1.7}
& \meanstd{66.4}{1.1}
& \meanstd{82.4}{2.5}
& \meanstd{69.8}{1.8} \\

\textit{w/o Sign Alignment}
& \bfseries\boldmath\meanstd{93.1}{3.1}
& \meanstd{41.3}{6.9}
& \meanstd{79.9}{2.3}
& \meanstd{71.0}{5.9}
& \meanstd{72.8}{8.4}
& \meanstd{50.8}{1.4}
& \meanstd{71.4}{0.5}
& \meanstd{81.5}{2.4}
& \meanstd{71.6}{1.5} \\

\textbf{OCSD (ours)}
& \meanstd{91.9}{1.4}
& \bfseries\boldmath\meanstd{77.2}{4.4}
& \bfseries\boldmath\meanstd{90.2}{4.6}
& \bfseries\boldmath\meanstd{84.7}{4.4}
& \meanstd{80.9}{4.3}
& \meanstd{63.8}{0.9}
& \bfseries\boldmath\meanstd{82.8}{0.7}
& \bfseries\boldmath\meanstd{86.9}{1.7}
& \bfseries\boldmath\meanstd{73.7}{1.6} \\

\bottomrule
\end{tabular*}

\caption{
Detailed ablation results with Qwen3-4B on ALFWorld and WebShop.
Results are reported as mean $\pm$ standard deviation over three random seeds. The best result in each column is shown in bold.
}
\label{tab:ablation_details}
\end{table*}

Table~\ref{tab:ablation_details} presents the complete results for the four
OCSD ablations, including subtask-level success rates on ALFWorld and both
score and success rate on WebShop. All variants use Qwen3-4B under the same
training configuration, differing only in whether they remove the
Observation-Ablated Teacher, replace high-NLL step selection with random
selection, apply residual calibration to all interaction steps, or remove
sign alignment.

\section{Prompts}
\label{app:prompts}

This section provides the benchmark-specific interaction prompts used for
rollout generation and evaluation, together with the replay-scoring prompt
templates used by OCSD. All methods use the same interaction prompts for
environment rollouts and evaluation. OCSD additionally constructs the Full
and Observation-Ablated prompt variants only for post-rollout replay scoring.

\subsection{Interaction Prompts}
\label{app:interaction_prompts}

The interaction prompt at each step contains the task instruction, interaction
history, current observation, and benchmark-specific action interface. The
model is instructed to place its intermediate reasoning within
\texttt{<reason>} and \texttt{</reason>} tags before producing an executable
action in the format required by the corresponding benchmark. Qwen3's native
thinking mode is disabled to avoid duplicated reasoning delimiters.

The interaction prompt templates for ALFWorld, WebShop, and Search-QA are
shown in Tables~\ref{tab:alfworld_prompt}, \ref{tab:webshop_prompt},
and~\ref{tab:searchqa_prompt}, respectively.

\subsection{Replay-Scoring Prompts}
\label{app:replay_prompts}

The replay-scoring prompts retain the original task instruction, interaction
history, current observation, and output-format instruction, and append the
corresponding replay evidence. The Full and Observation-Ablated prompts use
the same field order and future-action scaffold, differing only in whether the
actual future observations are included.

Tables~\ref{tab:full_replay_prompt}
and~\ref{tab:ablated_replay_prompt} show the Full and Observation-Ablated
prompt templates for the same ALFWorld interaction step. In the latter, each
future-observation field is replaced with the fixed phrase
\texttt{Observation: not provided}. WebShop and Search-QA use the same
construction with their respective interaction and action formats.

\begin{table*}[tbp]
\centering
\small

\begin{tabular}{
|p{\dimexpr\textwidth-2\tabcolsep-2\arrayrulewidth\relax}|
}
\hline

\rule[-0.6ex]{0pt}{2.8ex}
\textbf{Interaction Prompt on ALFWorld} \\

\hline

You are an expert agent operating in the ALFRED Embodied Environment. Your task is to: \{task\_description\}\newline
Prior to this step, you have already taken \{step\_count\} step(s). Below are the most recent \{history\_length\} observations and the corresponding actions you took: \{action\_history\}\newline
You are now at step \{current\_step\} and your current observation is: \{current\_observation\}\newline
Your admissible actions of the current situation are: [\{admissible\_actions\}].\newline
\newline
Now it's your turn to take an action.\newline
You should first reason step-by-step about the current situation. This reasoning process MUST be enclosed within \texttt{<reason> </reason>} tags.\newline
Once you've finished your reasoning, you should choose an admissible action for current step and present it within \texttt{<action> </action>} tags. \\

\hline
\end{tabular}

\caption{Prompt template used by OCSD for the ALFWorld task environment.}
\label{tab:alfworld_prompt}
\end{table*}

\begin{table*}[tbp]
\centering
\small

\begin{tabular}{
|p{\dimexpr\textwidth-2\tabcolsep-2\arrayrulewidth\relax}|
}
\hline

\rule[-0.6ex]{0pt}{2.8ex}
\textbf{Interaction Prompt on WebShop} \\

\hline

You are an expert autonomous agent operating in the WebShop e-commerce environment.\newline
Your task is to: \{task\_description\}.\newline
Prior to this step, you have already taken \{step\_count\} step(s). Below are the most recent \{history\_length\} observations and the corresponding actions you took: \{action\_history\}\newline
You are now at step \{current\_step\} and your current observation is: \{current\_observation\}.\newline
Your admissible actions of the current situation are: [\{available\_actions\}].\newline
\newline
Now it's your turn to take one action for the current step.\newline
You should first reason step-by-step about the current situation, then think carefully which admissible action best advances the shopping goal. This reasoning process MUST be enclosed within \texttt{<reason> </reason>} tags.\newline
Once you've finished your reasoning, you should choose an admissible action for current step and present it within \texttt{<action> </action>} tags. \\

\hline
\end{tabular}

\caption{Prompt template used by OCSD for the WebShop task environment.}
\label{tab:webshop_prompt}
\end{table*}

\begin{table*}[tbp]
\centering
\small

\begin{tabular}{
|p{\dimexpr\textwidth-2\tabcolsep-2\arrayrulewidth\relax}|
}
\hline

\rule[-0.6ex]{0pt}{2.8ex}
\textbf{Interaction Prompt on Search-QA} \\

\hline

You are an expert agent tasked with answering the given question step-by-step.\newline
Your question: \{task\_description\}\newline
\newline
Prior to this step, you have already taken \{step\_count\} step(s). Below is the interaction history where \texttt{<search> </search>} wrapped your past search queries and \texttt{<information> </information>} wrapped the corresponding search results returned by the external search engine. History:\newline
\{memory\_context\}\newline
\newline
Now it's your turn to respond for the current step.\newline
You should first conduct reasoning process. This process MUST be enclosed within \texttt{<reason> </reason>} tags.\newline
After completing your reasoning, choose only one of the following actions (do not perform both):\newline
(1) If you find you lack some knowledge, you can call a search engine to get more external information using format: \texttt{<search> your query </search>}.\newline
(2) If you have enough knowledge to answer the question confidently, provide your final answer within \texttt{<answer> </answer>} tags, without detailed illustrations. For example, \texttt{<answer>Beijing</answer>}. \\

\hline
\end{tabular}

\caption{Prompt template used by OCSD for the Search-QA task environment.}
\label{tab:searchqa_prompt}
\end{table*}

\begin{table*}[tbp]
\centering
\small

\begin{tabular}{
|p{\dimexpr\textwidth-2\tabcolsep-2\arrayrulewidth\relax}|
}
\hline

\rule[-0.6ex]{0pt}{2.8ex}
\textbf{Prompt of Full Replay-Evidence Scoring on ALFWorld} \\

\hline

You are an expert agent operating in the ALFRED Embodied Environment. Your task is to: \{task\_description\}\newline
Prior to this step, you have already taken \{step\_count\} step(s). Below are the most recent \{history\_length\} observations and the corresponding actions you took: \{action\_history\}\newline
You are now at step \{current\_step\} and your current observation is: \{current\_observation\}\newline
Your admissible actions of the current situation are: [\{admissible\_actions\}].\newline
\newline
Now it's your turn to take an action.\newline
You should first reason step-by-step about the current situation. This reasoning process MUST be enclosed within \texttt{<reason> </reason>} tags.\newline
Once you've finished your reasoning, you should choose an admissible action for the current step and present it within \texttt{<action> </action>} tags.\newline
\newline
Future evidence:\newline
After current action:\newline
Observation: \{future\_observation\_1\}\newline
\newline
Next action:\{future\_action\_schema\}\newline
\newline
After next action:\newline
Observation: \{future\_observation\_2\} \\

\hline
\end{tabular}

\caption{Template of the Full replay-evidence scoring prompt for the ALFWorld task environment.}
\label{tab:full_replay_prompt}
\end{table*}

\begin{table*}[tbp]
\centering
\small

\begin{tabular}{
|p{\dimexpr\textwidth-2\tabcolsep-2\arrayrulewidth\relax}|
}
\hline

\rule[-0.6ex]{0pt}{2.8ex}
\textbf{Prompt of Observation-Ablated Replay-Evidence Scoring on ALFWorld} \\

\hline

You are an expert agent operating in the ALFRED Embodied Environment. Your task is to: \{task\_description\}\newline
Prior to this step, you have already taken \{step\_count\} step(s). Below are the most recent \{history\_length\} observations and the corresponding actions you took: \{action\_history\}\newline
You are now at step \{current\_step\} and your current observation is: \{current\_observation\}\newline
Your admissible actions of the current situation are: [\{admissible\_actions\}].\newline
\newline
Now it's your turn to take an action.\newline
You should first reason step-by-step about the current situation. This reasoning process MUST be enclosed within \texttt{<reason> </reason>} tags.\newline
Once you've finished your reasoning, you should choose an admissible action for the current step and present it within \texttt{<action> </action>} tags.\newline
\newline
Future evidence:\newline
After current action:\newline
\texttt{Observation: not provided}\newline
\newline
Next action:\{future\_action\_schema\}\newline
\newline
After next action:\newline
\texttt{Observation: not provided} \\

\hline
\end{tabular}

\caption{Template of the Observation-Ablated replay-evidence scoring prompt for the ALFWorld task environment.}
\label{tab:ablated_replay_prompt}
\end{table*}

\end{document}